\documentclass[lettersize,journal]{IEEEtran}
\usepackage{amsmath,amsfonts}
\usepackage{algorithmic}
\usepackage{algorithm}
\usepackage{array}
\usepackage{textcomp}
\usepackage{stfloats}
\usepackage{url}
\usepackage{verbatim}
\usepackage{graphicx}
\usepackage{cite}
\usepackage{mathtools}
\usepackage{subfloat}
\usepackage{multirow}
\usepackage[dvipsnames]{color,xcolor}
\usepackage{subfigure}
\usepackage{booktabs}
\usepackage{bbding}

\newcommand{\yep}[1]{\textcolor{black}{#1}}

\newcommand{\review}[1]{\textcolor{black}{#1}}
\graphicspath{{./output/}}
\begin{document}

\title{Exploring Multi-Timestep Multi-Stage Diffusion Features for Hyperspectral Image Classification}

\author{Jingyi Zhou$^*$, 
Jiamu Sheng$^*$, 
Peng Ye,
Jiayuan Fan,~\IEEEmembership{Member,~IEEE,} 
Tong He, \\
Bin Wang,~\IEEEmembership{Senior Member,~IEEE,}
and Tao Chen,~\IEEEmembership{Senior Member,~IEEE}\thanks{This work is supported by National Key R\&D Program of China (2022ZD0160300), National Natural Science Foundation of China (No. 62101137 and 62071127),  Shanghai Natural Science Foundation (No. 23ZR1402900). The computations in this research were performed using the CFFF platform of Fudan University. \textit{(Corresponding author: Jiayuan Fan.)}}
\thanks{Jingyi Zhou, Peng Ye, Bin Wang and Tao Chen are with School of Information Science and Technology, Fudan University, Shanghai 200433, China (e-mail: zhoujingyi19@fudan.edu.cn; eetchen@fudan.edu.cn).}
\thanks{Jiamu Sheng and Jiayuan Fan are with the Academy for Engineering and Technology, Fudan University, Shanghai 200433, China (e-mail: jmsheng22@m.fudan.edu.cn; jyfan@fudan.edu.cn).}% <-this % stops a space
\thanks{Tong He is with the Shanghai Artificial Intelligence Laboratory, Shanghai 200232, China.}\thanks{$^*$Jingyi Zhou and Jiamu Sheng contributed equally to this work.}} %Jiayuan Fan is the corresponding author.
%}% <-this % stops a space


\maketitle

\maketitle

\vspace{-3mm}
\begin{abstract}

The effectiveness of spectral-spatial feature learning 
% determines the capacity of 
is crucial for the hyperspectral image (HSI) classification task. Diffusion models, as a new class of groundbreaking generative models, have the ability to learn both contextual semantics and textual details from the distinct timestep dimension, enabling the modeling of complex spectral-spatial relations in HSIs. However, existing diffusion-based HSI classification methods only utilize manually selected single-timestep single-stage features, limiting the full exploration and exploitation of rich contextual semantics and textual information hidden in the diffusion model. To address this issue, we propose a novel diffusion-based feature learning framework that explores Multi-Timestep Multi-Stage Diffusion features for HSI classification for the first time, called MTMSD. Specifically, the diffusion model is first pretrained with unlabeled HSI patches to mine the connotation of unlabeled data, and then is used to extract the multi-timestep multi-stage diffusion features. To effectively and efficiently leverage multi-timestep multi-stage features,
% to integrate contextual semantics and textual features, 
two strategies are further developed. 
% for MTMSD feature exploration. 
One strategy is class \& timestep-oriented multi-stage feature purification module with the inter-class and inter-timestep prior for reducing the redundancy of multi-stage features and alleviating memory constraints. The other one is selective timestep feature fusion module with the guidance of global features to adaptively select different timestep features for integrating texture and semantics. Both strategies facilitate the generality and adaptability of the MTMSD framework for diverse patterns of different HSI data. Extensive experiments are conducted on \review{four} public HSI datasets, and the results demonstrate that our method outperforms state-of-the-art methods for HSI classification, especially on the challenging Houston 2018 dataset. 
The codes are available at https://github.com/zjyaccount/MTMSD.
\end{abstract}


\vspace{-0.6em}
\section{Introduction}
%\IEEEPARstart{H}{YPERSPECTRAL} image (HSI) classification is an important research topic in the area of remote sensing (RS) \cite{fan2017superpixeltcsvt,xie2020multiscaletcsvt,2022tcsvt2,2023tcsvt1}. HSIs, distinguished by their detailed spectral information, enable effective discrimination of objects of interest by capturing more subtle discrepancies from the contiguous shape of the spectral signatures associated with their pixels \cite{zhang2018fasttcsvt}. 
%Compared to RGB and multi-spectral data, HSI allows for finer and more accurate detection and recognition of materials on the earth's surface \cite{xu2017multisource}. 
%However, the high spectral variability makes discriminative information \jy{extraction} from such data difficult \cite{hong2018unmixing}. \jy{Since HSI classification aims to distinguish each pixel's category in hyperspectral data using dense electromagnetic spectral information \cite{li2019deep}, a large number of handcrafted feature-based methods have been designed for HSI classification \cite{bandos2009classification,ghamisi2017advances, fauvel2008spectral, hong2020invariant, peng2017robust}. while they are designed manually and limited in spectral-spatial feature extraction. 

\IEEEPARstart{H}{YPERSPECTRAL} image (HSI) classification plays a crucial role in remote sensing, as it aims to distinguish each pixel's category in hyperspectral data by using dense and detailed electromagnetic spectral information~\cite{li2019deep, zhang2022artificial}. HSI classification has a wide range of applications in environmental monitoring \cite{obermeier2019grassland}, resource management \cite{wu2020resourse}, agriculture disaster response \cite{hsieh2020agriculture}, military defense \cite{shimoni2019military}, etc.

The extraction of spectral-spatial features from HSIs plays a significant role in HSI classification tasks. Initially, machine learning-based feature extraction methods have been developed~\cite{bandos2009classification,ghamisi2017advances, fauvel2008spectral, hong2020invariant, peng2017robust}. However, these approaches rely on manual feature engineering, resulting in limited extraction of discriminative information from the highly variable spectral data~\cite{hong2018unmixing}. In recent years, the rapid advancement of deep learning has paved the way for neural network-based approaches for feature extraction, specifically leveraging convolutional networks (CNNs)~\cite{yang20182dcnn,zhong2018ssrn,yu2021feedback}  and transformer models~\cite{hong2021spectralformer, sun2022ssftt, mei2022gaht, wuke2023hyperspectral}. These neural network-based methods excel at automatically learning valuable spectral-spatial features
% discovering potential relations and valuable knowledge
from labeled HSI data, thus achieving promising results in HSI classification. Further, unsupervised methods such as \cite{mei2019unsupervised3dcae, zhang2022unsupervisedumsdfl,zhumingzhen} have been designed to excavate spectral-spatial features from unlabeled HSIs. Typically, these methods employ an encoder-decoder network trained in an unsupervised manner for HSI reconstruction, thus extracting spectral-spatial features from unlabeled HSI data. %Leveraging unsupervised feature extraction can overcome the limitations of supervised methods, particularly the scarcity of labeled samples in HSI datasets.
Leveraging unsupervised feature learning enables deep mining of large amounts of unlabeled regions in HSI, which contain a wealth of label-agnostic information, thereby promoting the HSI classification task.
\begin{figure}[t]
    \centering
    \includegraphics[width=8.5cm]{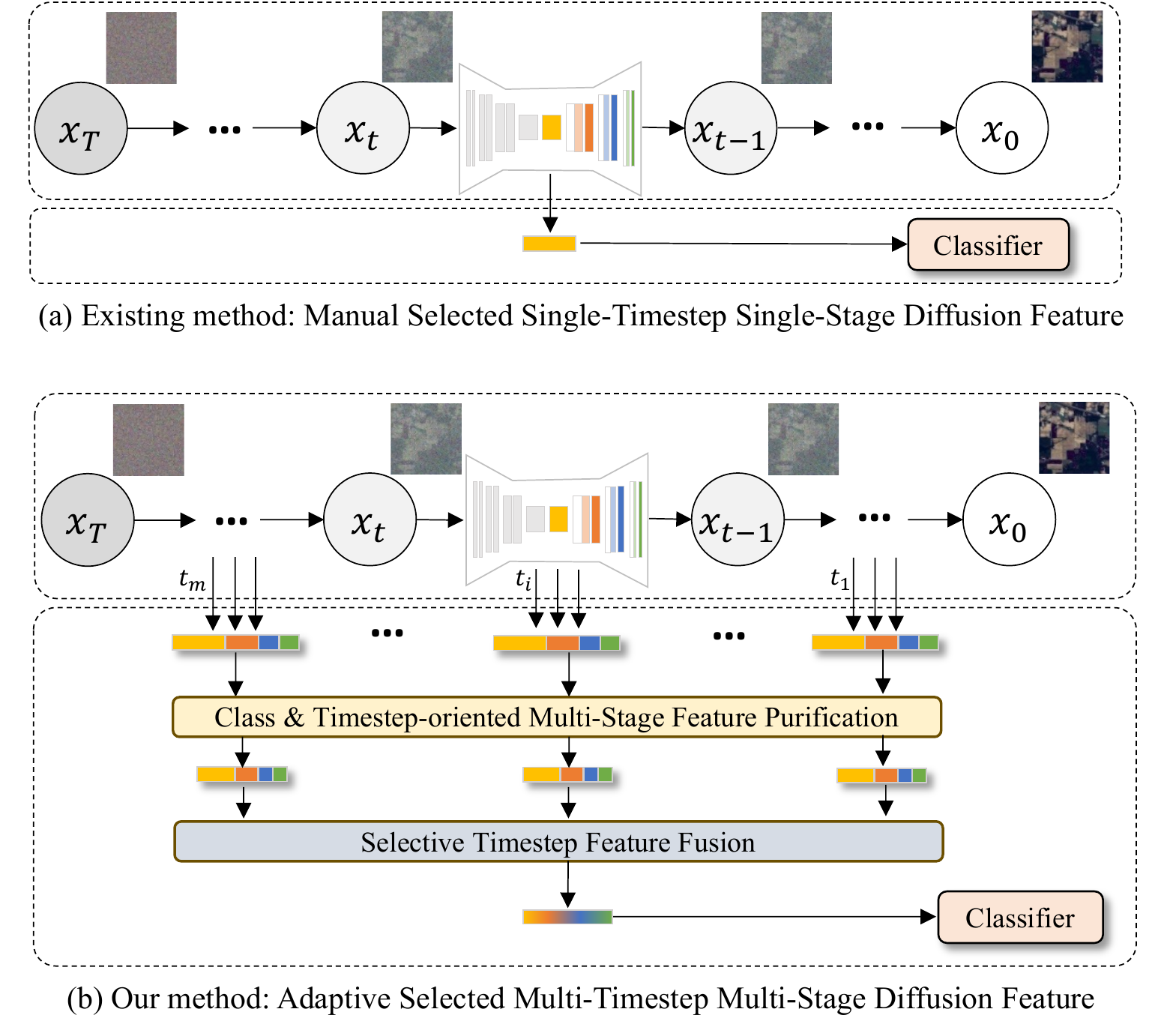}
     \caption{Overview of existing diffusion-based feature learning frameworks for the HSI classification task. Our method can fully explore and exploit rich contextual semantics and textual features hidden in the diffusion model.}
     \vspace{-1em}
    \label{fig1}
\end{figure}

Recently, diffusion models~\cite{ho2020ddpm, sohl2015diffusion, nichol2021improvedddpm} have emerged as powerful models with superior performance in generation and reconstruction tasks. These models have also been explored in various computer vision tasks such as semantic segmentation~\cite{Baranchuk2021LabelEfficientSS, wolleb2022diffusionmedical, wang2023segrefiner}. Distinguished from traditional deep neural networks, diffusion models employ a stepwise reverse denoising process, formulated as an iterative optimization procedure optimized by Langevin dynamics~\cite{welling2011langevin}. This approach introduces a multitude of degrees of freedom for feature learning and predicts additional information conditioned on the given noise-corrupted data at each timestep. Consequently, diffusion models can implicitly capture both high-level and low-level visual concepts, facilitating better generalization and modeling of complex spectral-spatial relations~\cite{zhao2023unleashing, song2019sgm}. Therefore, one recent work has proposed using diffusion models for HSI classification~\cite{spectraldiff}, which employs unsupervised feature learning to extract diffusion features. However, the diffusion features used in~\cite{spectraldiff} are extracted solely from a single timestep and a single stage of the denoising U-Net, and these selections about the timestep and stage are manually determined based on extensive experimentation with each dataset, as shown in Fig.~\ref{fig1}. Firstly, relying on single-timestep features from a single U-Net layer inevitably results in the loss of abundant spectral-spatial information and limits the effectiveness of modeling spectral-spatial relations. Secondly, the manual selection approach lacks generality and adaptability to diverse HSI datasets exhibiting specific spectral representation patterns.

% Nonetheless, 
Naturally, it is crucial to explore how to exploit abundant multi-timestep multi-stage features extracted from the whole stage of denoising U-Net more effectively and efficiently. First, multi-timestep features extracted from diffusion models are diverse and focus on different information. The shallow timestep-wise features are more informative for textual details, while deeper ones are more concerned about high-level semantics and global information \cite{ho2020ddpm, choi2022diffusionperception}. For modeling complex spectral-spatial relations, textual features depict the spatial distribution and changing patterns of ground objects, and contextual semantics represent the spectral attributes and the content of ground objects. Thus, leveraging multi-timestep features can integrate both contextual semantics and textual features to build comprehensive spectral-spatial representation for better performance.
Second, multi-stage features, as compared to single-stage features, encapsulate a rich hierarchy of information that contains richer semantic and reconstruction characteristics, facilitating the modeling of spectral-spatial relations. Moreover, \cite{spectraldiff} also demonstrates that various stage features improve classification performance with different degrees across various datasets.
%When applying diffusion models to unsupervised feature learning of HSI data, it not only learns both high-level and low-level features but also extracts spectral-spatial features in the dimension of timestep $t$, as shown in Fig.~\ref{1b}. 
% Compared to common unsupervised feature learning methods, diffusion models can capture spectral-spatial features in the dimension of timestep $t$, as shown in Fig.~\ref{1b}, enriching the representation capability of the features. 
%\yep{These substantial timestep-wise features captured from diffusion models are abundant and informative, so manually selecting timestep-wise features for HSI classification is difficult to characterize spectral-spatial features effectively and robustly. } 
Although multi-timestep multi-stage features are desired, HSIs from various datasets acquired by different sensors exhibit distinct spectral-spatial characteristics, leading to variations in the spectral-spatial representations. Moreover, for different regions in the same HSI, the emphasis on texture and semantics varies, leading to distinct preferences in the selection of timestep $t$. 
Meanwhile, multi-stage features are numerous and comprise both redundant semantics and reconstruction information along the channel dimension, posing a challenge in terms of memory constraints when training and inferring on large datasets. 

In view of these, we propose a novel diffusion-based feature learning framework that explores \textbf{M}ulti-\textbf{T}imestep \textbf{M}ulti-\textbf{S}tage \textbf{D}iffusion features for HSI classification for the first time, named \textbf{MTMSD}.
%\yep{Leveraging two significant properties of diffusion models,} MTMSD can implicitly learn both high-level and low-level features to more effectively model the complex nonlinear relations between the spectral bands of HSIs, and represent the spatial structure and distribution. 
More specifically, the proposed framework first pretrain a diffusion model with unlabeled HSI patches for diffusion feature learning, mining the connotation of unlabeled data that reveals the complex spectral-spatial dependencies. 
Then, we extract multi-timestep multi-stage diffusion features from the pretrained denoising U-Net decoder and construct the timestep-wise center and global feature bank by center extraction and average pooling. 
%Global diffusion features are employed to provide global semantic information to the corresponding central diffusion features for modeling spatial distribution. 
After that, two strategies are further developed in MTMSD to leverage multi-timestep multi-stage diffusion features effectively and efficiently. First, to reduce the redundancy of multi-stage features and maintain efficiency, we propose to perform class \& timestep-oriented multi-stage feature purification on multi-stage features in the timestep-wise center feature bank with the inter-class and inter-timestep prior.
%which contains various informative features from different timestep $t$}. 
Second, to effectively harness multi-timestep features and softly learn the proper timestep-wise feature combination for different datasets, we propose the selective timestep feature fusion module. This module is designed to adaptively select different timestep center features with the guidance of related global features,
% to effectively harness multi-timestep features and softly learn the proper timestep-wise feature combination for different datasets.
% for modeling spatial distribution, 
and fuse them for multi-timestep multi-stage selective representations that integrate contextual semantics and textual features to model comprehensive spectral-spatial relations. 
Ultimately, an ensemble of linear classifiers is employed for accurate HSI classification.
%In order to efficiently exploit features from the timestep-wise feature bank and softly learn the proper timestep-wise feature combination for different HSI datasets, the dynamic feature fusion module is designed to adaptively fuse hierarchical features from different timesteps and generate multi-timestep representations with rich spectral-spatial information. Finally, an ensemble of linear classifiers is applied to these dynamic representations to perform HSI classification.

To summarize, our contributions are listed as follows.
\begin{itemize}
%\item[1)] As a pioneer work, we introduce diffusion models to HSI classification and propose the unsupervised diffusion-based spectral-spatial feature learning framework, called MTMSD. We first pretrain a diffusion model for unsupervised feature learning, which can automatically learn the label-agnostic prior of HSI data. Then we extract hierarchical features from pretrained diffusion model and learn multi-timestep representations, which can implicitly learn both high-level and low-level features to model the complex nonlinear relations between the spectral bands of HSIs, and represent the spatial structure and distribution. 
%\item[1)] As a pioneer work, we introduce diffusion models to HSI classification and propose a diffusion-based unsupervised feature learning framework called MTMSD, which can implicitly learn both high-level and low-level features to model complex spectral-spatial relations.
%\item[1)] We propose a diffusion-based multi-timestep spectral-spatial feature learning framework, called MTMSD. Compared to existing methods, it can implicitly learn multi-timestep diffusion features by iterative denoising to model complex spectral-spatial relations. 
%\item [1)] For modeling complex spectral-spatial relations, we delve into the role of diffusion models in unsupervised feature learning and discover that the timestep in diffusion models serves as a valuable feature dimension for HSI classification.

\item [1)] For modeling complex spectral-spatial relations, we propose a novel diffusion-based framework that explores multi-timestep multi-stage diffusion features for HSI classification. To the best of our knowledge, this is the first work to learn and exploit multi-timestep multi-stage diffusion features for diffusion-based HSI classification.

\item[2)] We design the novel class \& timestep-oriented multi-stage feature purification module. It adaptively selects significant channels of multi-stage diffusion features from both inter-class and inter-timestep aspects to reduce redundant information and maintain computational efficiency.

%To leverage abundant purified timestep-wise features effectively and efficiently, 
\item[3)] We propose to perform selective timestep feature fusion on multi-timestep diffusion features. This module allows each labeled patch of different datasets to adaptively select different timestep center features with the guidance of related global features, and fuse them for comprehensive multi-timestep multi-stage selective representations that integrate contextual semantics and textual information.
%This process dynamically learns the soft timestep-wise feature combination for each labeled patch of different HSI datasets and constructs comprehensive multi-timestep representations.

%\item[3)] Compared with several CNN-based and transformer-based supervised methods and some state-of-the-art unsupervised feature learning methods, our method is more effective for HSI classification. Experimental results demonstrate that the proposed method achieves superior classification accuracy on three HSI datasets.
\item[4)] Compared with several state-of-the-art HSI classification methods, experimental results demonstrate that our proposed method achieves significant classification accuracy on \review{four} public HSI datasets, especially on the challenging Houston 2018 dataset.
\end{itemize}

The remainder of this paper is organized as follows. Section II describes related work. In Section III, our proposed MTMSD is introduced in detail. Section IV conducts extensive experiments on \review{four} HSI datasets to demonstrate the effectiveness of the proposed method. Finally, some conclusions are drawn in Section V.

\section{Related Work}
\subsection{HSI Classification}
HSI classification is an important research topic in the area of remote sensing. Since HSI classification aims to distinguish each pixel's category in hyperspectral data using dense electromagnetic spectral information \cite{li2019deep}, a large number of handcrafted feature-based methods have been designed for HSI classification \cite{bandos2009classification,ghamisi2017advances, fauvel2008spectral, hong2020invariant, peng2017robust}.
Several works adopt morphological profiles (MPs) for manually extracting spectral-spatial features from HSIs. They achieve good classification results using MPs as input vectors with a support vector machine classifier. 
Subspace-based learning, such as sparse representation and manifold learning, is another common feature extraction strategy for HSI classification. These methods transform the high-dimensional original space using a low-dimensional subspace representation to learn spectral-spatial information. 

Due to the remarkable breakthroughs achieved by deep learning in various computer vision tasks, many progressive deep learning-based networks have been widely utilized for HSI classification methods.
%e.g., autoencoders (AEs) \cite{mei2019unsupervised3dcae}, deep belief networks (DBNs) \cite{chen2015dbn} and recurrent neural networks (RNNs) \cite{hang2019rnn}. 
Among these, CNNs draw significant attention with their feature extraction capability to extract spatially structural information and locally contextual information and become mainstream in HSI classification \cite{yang20182dcnn,zhong2018ssrn,yu2021feedback}. 
Based on the spectral and spatial attention modules, Zhu \emph{et al.} \cite{zhong2018ssrn} embed a residual block into a sequential spectral-spatial feature learning network. This architecture not only mitigates the risk of overfitting but also enhances classification performance.
However, CNNs have the challenge of modeling long-term spectral information dependencies. To address this defect, researchers explore the value of the transformer and widely leverage it in HSI classification~\cite{hong2021spectralformer, sun2022ssftt, mei2022gaht, wuke2023hyperspectral}. 
%As the vision transformer rises in the field of computer vision, researchers begin to explore the value of the transformer in HSI classification and find it outperforms CNN-based methods in solving the problem of long-term spectral information dependencies \cite{hong2021spectralformer, sun2022ssftt, mei2022gaht, wuke2023hyperspectral}. 
Hong \emph{et al.} \cite{hong2021spectralformer} consider the spectral sequence of neighboring bands and design a pure transformer-based SpectralFormer (SF) backbone network, representing sequence attributes of spectral signatures. Sun \emph{et al.} \cite{sun2022ssftt} extract high-level semantic features by introducing a Gaussian weighted token module into the transformer architecture, achieving promising performance in both classification accuracy and computational complexity.

In addition to supervised feature learning, unsupervised feature learning 
% is a feature extraction paradigm that 
aims to learn feature representations from the input data without any annotated information,
% . Leveraging unsupervised feature learning on unlabeled HSI patches is 
providing a solution to the limited labeled samples of HSI datasets. Typically, the commonly used unsupervised feature learning methods in HSI classification are based on the encoder–decoder paradigm, where an autoencoder-like network encodes the input HSI patches into a purified feature and then reconstructs the feature to initial HSI data by a decoder network. Mou \emph{et al.} \cite{mou2017unsupervisedconvdeconv} first design a fully 2D Conv–Deconv network in an end-to-end manner for unsupervised feature learning of HSI classification. Similarly, Mei \emph{et al.} \cite{mei2019unsupervised3dcae} design a 3D convolutional autoencoder (3D-CAE) for unsupervised feature learning of HSI classification. To alleviate the insufficiency of geometric representation and exploit the multi-scale features, Zhang \emph{et al.} \cite{zhang2022unsupervisedumsdfl} design a multi-scale CNN-based unsupervised feature learning framework, with two branches of decoder and clustering optimized by the error feedback of image reconstruction and pseudo-label classification. 
%In the aforementioned approaches, the feature representations learned from the unlabeled HSI patches contain more spectral-spatial information than supervised methods; however, they mainly focus on low-level visual concepts and ignore high-level feature extraction for HSI classification, which is solved in our methods.

More recently, with the rise of diffusion models, one recent work propose a diffusion-based HSI classification method. Chen \emph{et al.} propose SpectralDiff \cite{spectraldiff} that extracts the spectral-spatial diffusion features from spectral–spatial denoising network and directly feeds them into the attention-based classification network for classification. However, they only use a single timestep feature from a single stage of the denoising U-Net, which is manually selected according to extensive experiments on each dataset. This results in a lack of information to model spectral-spatial relations and poor robustness to diverse HSI datasets with diverse spectral characteristics. In our work, we propose a novel diffusion-based HSI classification framework that explores multi-timestep multi-stage diffusion features. Through multi-stage feature purification and selective timestep fusion, 
our MTMSD enables adaptive integration of both contextual semantics and textual
features to model complex spectral-spatial relations and generality for diverse patterns of different HSI data.
\subsection{Diffusion Models}
Diffusion models are a class of probabilistic generative models that progressively inject a standard Gaussian noise, then learn a model to reverse this process for sample generation \cite{ho2020ddpm, sohl2015diffusion, nichol2021improvedddpm}. Current research on diffusion models is mostly based on three formulations: denoising diffusion probabilistic models (DDPMs) \cite{ho2020ddpm}, score-based generative models \cite{song2019sgm}, and stochastic differential equations \cite{song2021score}. 
%Ho \emph{et al.} highlighted the equivalence of diffusion models and score matching, showing them to be two different perspectives on the gradual conversion of a simple known distribution into a target distribution via the iterative denoising process. 
Among them, DDPMs are the mainstream diffusion models, and a large number of recent works based on DDPMs have made DDPMs increasingly powerful in terms of generative quality and diversity over other generative models \cite{dhariwal2021beatgan}. Meanwhile, DDPMs have been widely used in several applications, including super-resolution \cite{superresolutiondiff}, inpainting \cite{lugmayr2022inpaint}, and point cloud generation \cite{luo2021pointcloud}. Recently,~\cite{baranchuk2021labelefficient} proposes a simple diffusion-based semantic segmentation approach that exploits 3-timestep multi-stage features with manually selected timesteps and also proves that diffusion features capture high-level semantic information for semantic segmentation. However, a fixed timestep set results in suboptimal and non-generic features for different datasets. Additionally, such a simple utilization of diffusion features for segmentation may lead to inefficiencies due to redundant information and ineffectiveness in building semantic representations from diffusion features. Differently, we focus on the HSI classification task, and dynamically select the timesteps and stages of multi-timestep multi-stage features ranging from low to high, integrating textural features and semantics to build comprehensive spectral-spatial representations.

\subsection{Deep Feature Selection}
Feature selection \cite{han2021dynamicsurvey}, an essential process in deep-learning-based computer vision, plays a pivotal role in improving model performance. The goal of feature selection is to retain informative and refined features from the original features, thereby reducing dimensionality and computational complexity. For different images, the important features vary, and feature selection assists models in adapting to data from diverse domains, extracting significant and refined features.
Early feature selection methods select features with input-dependent soft attention. Hu \emph{et al.} \cite{hu2018senet} introduce a channel attention module named SENet to adaptively recalibrate channel-wise features by exploiting the inter-channel relations. Similar to channel attention, spatial attention methods GENet \cite{hu2018genet} is designed to enhance a network's capacity for context information modeling via spatial masks. Building on these, Woo \emph{et al.} \cite{woo2018cbam} combine both channel and spatial attention, introducing the CBAM method. Additionally, feature selection on different kernel features is also a self-adaptive and effective mechanism. Li \emph{et al.} \cite{li2019sknet} propose SKNet to select the features extracted from different convolutional kernels using softmax attention along the channel dimension. 
%Zhang \emph{et al.} \cite{zhang2022resnest} design ResNeSt by partitioning the input feature map into several groups, and selecting the group features similar to SKNet. 
Different from the above methods, %previously introduced, 
our framework is the first to use feature selection on multi-timestep multi-stage features for HSI classification. In detail, MTMSD creatively employs inter-class and inter-timestep priors for multi-stage feature purification in the channel dimension, and concurrently conducts feature selection on multi-timestep features with the guidance of global information.

\begin{figure*}[t!]
    \centering
    \includegraphics[width=18cm]{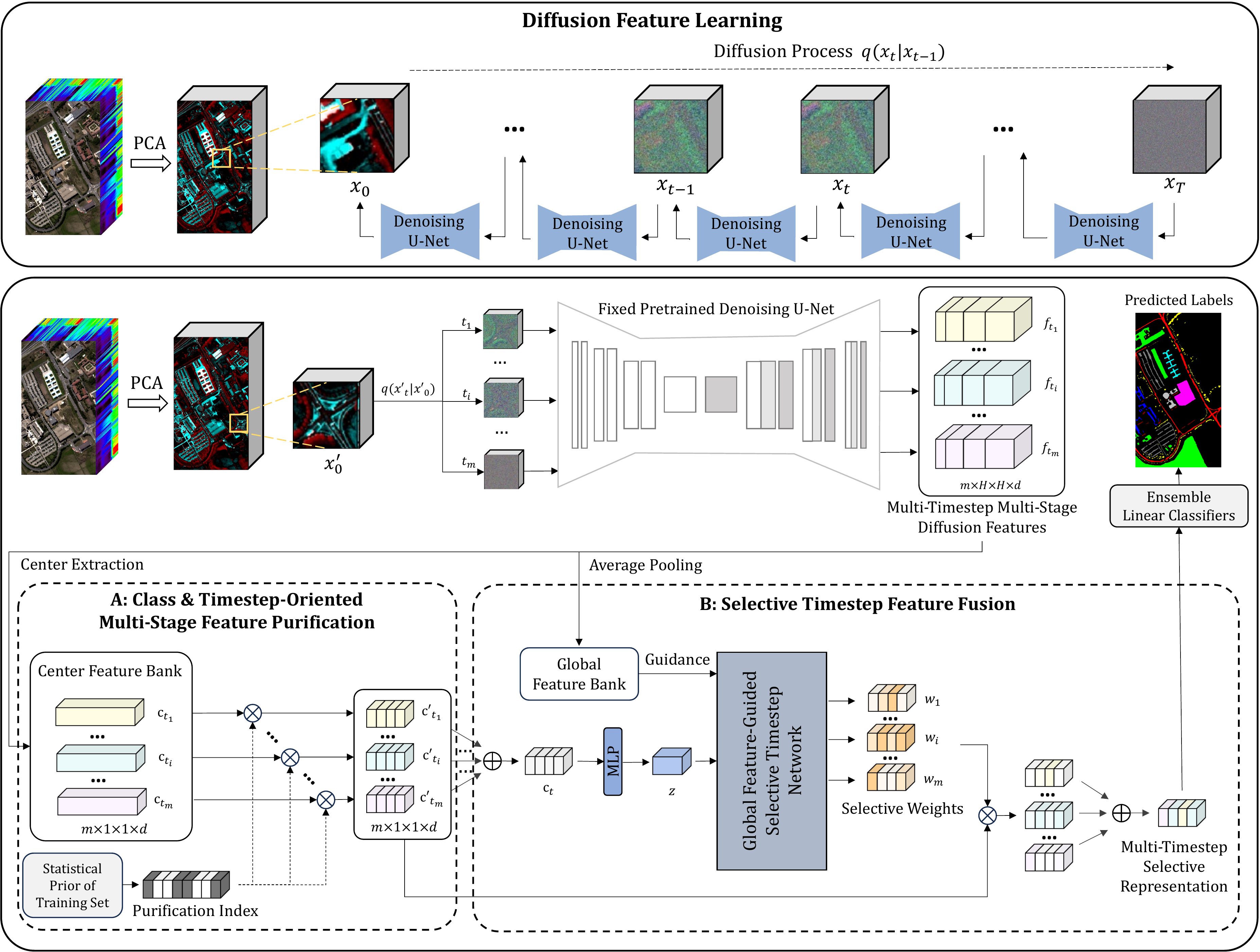}
     \caption{Overview of our proposed MTMSD. The method consists of two steps. Step 1: We pretrain the DDPM with HSI patches in an unsupervised manner for diffusion feature learning. Step 2: We extract multi-timestep multi-stage diffusion features from the pretrained denoising U-Net decoder and construct the timestep-wise center and global feature bank by center extraction and average pooling. To effectively and efficiently leverage multi-timestep multi-stage features, we first perform class \& timestep-oriented multi-stage purification on multi-stage features in the timestep-wise center feature bank, and then, we perform selective timestep feature fusion with global-feature guidance on the purified timestep-wise center feature bank. Classification is performed through an ensemble of lightweight classifiers.}
     \vspace{-0.5em}
    \label{fig:method}
\end{figure*}

\section{Method}
%Our proposed MTMSD develops two strategies, class \& timestep-oriented multi-stage purification module and selective timestep feature fusion module, to explore multi-timestep multi-stage diffusion features effectively and efficiently for modeling spectral-spatial relations comprehensively. The framework is shown in Fig.~\ref{fig:method}. In this section, we introduce our proposed MTMSD in detail.
Our proposed MTMSD is a novel diffusion-based feature learning framework and aims to explore multi-timestep multi-stage diffusion features effectively and efficiently for modeling spectral-spatial relations comprehensively. The framework is shown in Fig.~\ref{fig:method}. In the following section, we introduce the proposed MTMSD in detail.
%Our method consists of two key steps. The first step is to pretrain the DDPM with unlabeled patches for timestep-aware diffusion feature learning, which learns complex spectral-spatial distributions of HSI data from both semantic and textual aspects. After pretraining, we freeze the parameters of the pretrained DDPM. \yep{The second step of our MTMSD is a supervised learning process} aiming to fully extract and adaptively select the timestep-wise features from the pretrained denoising U-Net for HSI classification. We first extract multi-timestep diffusion features from all stages of the fixed pretrained denoising U-Net decoder and construct the timestep-wise center and global feature bank by center extraction and average pooling. Then, we design to perform class \& timestep-oriented diffusion feature purification on the timestep-wise center and global feature bank. To adaptively integrate contextual semantics and textual information in timestep-wise features, we perform a multi-timestep feature selective fusion process and generate multi-timestep multi-stage selective representations for classification. Finally, the representations are fed into an ensemble of lightweight classifiers to predict the label of the center pixel. 
%
%, including the unsupervised diffusion-based pretraining step as well as the supervised finetuning step with multi-timestep representation generation. 
%\subsection{Overview of Proposed MTMSD}

\subsection{Diffusion Feature Learning on unlabeled HSI}
%In order to learn complex spectral-spatial relations and label-agnostic information of HSI data, we pretrain a diffusion model in an unsupervised manner in the first step of our MTMSD. We first give a brief review of DDPM and introduce the basic principles to make our method easier to understand. Then, we explain our unsupervised feature learning procedure in detail through diffusion-based pretraining with HSI data.  
\subsubsection{A Brief Review of DDPM}
%Before formally introducing our method, it is necessary to introduce some prior knowledge of DDPM. 
\yep{DDPMs are a class of likelihood-based models that reconstruct the distribution of training data via an encoder-decoder denoising model. The denoising model is trained to remove noise from the training data destructed by Gaussian noises step-by-step.} These models consist of a forward noising process and a reverse denoising process. In the forward process, Gaussian noise is added to the original training data $x_0 \sim q\left(x_0\right)$ step by step over $T$ time steps, which follows the Markovian process:
\begin{equation}
\label{eq1}
q(x_t|x_{t-1})=\mathcal{N} (\sqrt{1-\beta_{t}}x_{t-1}, \beta_{t}I)
\end{equation}
where $\mathcal{N}(.)$ is a Gaussian distribution, and the Gaussian variances $\{\beta_{t}\}_{t=0}^{T}$ that determines the noise schedule are either be learned or scheduled. The above formulation leads that an arbitrary noisy sample $x_t$ for each timestep $t$ is obtained directly from $x_0$:
\begin{equation}
\label{eq2}
x_t = \sqrt{\overline{\alpha}_t}x_0 + \sqrt{(1 - \overline{\alpha}_t)}\epsilon,  
     \epsilon \sim \mathcal{N}(0,I)
\end{equation}
where $\alpha_t=1 - \beta_t$, and $\overline{\alpha}_t=\prod_{s=1}^t \alpha_s$. Then in the reverse process, DDPM also follows a Markovian process to denoise the noisy sample $x_T$ to $x_0$ step by step. Under large $T$ and small $\beta_t$, the reverse transitions probability is approximated as a Gaussian distribution and is predicted by a learned neural network as follows:
\begin{equation}
\label{eq3}
p_\theta(x_{t-1}|x_t)=\mathcal{N}(x_{t-1};\mu_\theta(x_t,t),\sigma_\theta(x_t,t))
\end{equation}
where the reverse process is re-parameterized by estimating $\mu_\theta(x_t,t)$ and $\sigma_\theta(x_t,t)$. $\sigma_\theta(x_t,t)$ is set to $\sigma_t^2 \mathbf{I}$, where $\sigma_t^2$ is not learned. In practice, rather than predicting $\mu_\theta(x_t,t)$ directly, predicting the noise $\epsilon$ in Eq.~\ref{eq2} via a U-Net works best, and the parameterization of $\mu_\theta(x_t,t)$ is derived as follows: 
\begin{equation}
    \label{eq4}
    \mu_\theta(x_t,t)=\frac{1}{\sqrt{\alpha_t}}(x_t-\frac{1-\alpha_t}{\sqrt{1-\overline{\alpha}_t}}\epsilon_\theta(x_t,t))
\end{equation}
The U-Net denoising model $\epsilon_\theta(x_t,t)$ is optimized by minimizing the following loss function:
\begin{equation}
\label{eq5}
\mathcal{L}(\theta)=E_{t,x_0,\epsilon}[(\epsilon-\epsilon_\theta(\sqrt{\overline{\alpha}_t}x_0)+\sqrt{1-\overline{\alpha}_t\epsilon},t)^2]
\end{equation}

\yep{In our work, improved DDPM~\cite{nichol2021improvedddpm} is adopted and has been proven to bring some improvements to the above DDPM. In detail, learned variances $\sigma_\theta(x_t,t)$ and an improved cosine noise schedule proposed in \cite{nichol2021improvedddpm} lead to enhanced distribution learning ability.}

%\begin{figure*}[thp]
%    \centering
%    \includegraphics[width=18cm]{tsne_v1.pdf}
%    \caption{2D t-SNE visualization of features from different timesteps on the Indian Pines dataset. (a) original spectral feature. (b) feature at $t=100$. (c) feature at $t=300$. (d) feature at $t=500$. (e) feature at $t=700$. (f) feature at $t=900$.}
%    \label{fig:tsne}
%\end{figure*}

%\subsubsection{DDPM Pretraining with Hyperspectral Data}
\subsubsection{Unsupervised Hyperspectral Diffusion Pretraining} 
Using the training skills and optimization objectives mentioned above, the DDPM for diffusion feature learning is trained with unlabeled hyperspectral data. Before training, the HSI data is pre-processed by principal components analysis (PCA) and random patch cropping operation. Then, given an unlabeled patch $x_0 \in \mathcal{R}^{H \times H \times D}$, where $H$ is the patch size, and $D$ is the number of PCA components, we gradually add Gaussian noise to the unlabeled HSI patch according to the cosine variance schedule $\{\beta_{t}\}_{t=0}^{T}$ in the diffusion process, where $T$ is the total number of the timestep. Then, in the reverse process, a denoising U-Net is trained to predict the noise added on $x_{t-1}$ taking noisy patch $x_t$ and timestep $t$ as inputs. And the sample $x_0$ can be obtained from the noise patch $x_t$ by the iterative denoising steps according to Eq.~\ref{eq3}. \yep{In each step of the training process, the timestep $t$ is randomly sampled from 0 to $T$.} The U-Net denoising model $\epsilon_\theta(x_t,t)$ is optimized by minimizing Eq.~\ref{eq5}. The parameters in the pretraining process, such as patch size, number of PCA components, and total pretraining steps, will be discussed in Sec IV. F.
\begin{figure*}[t!]
    \centering
    \includegraphics[width=18cm]{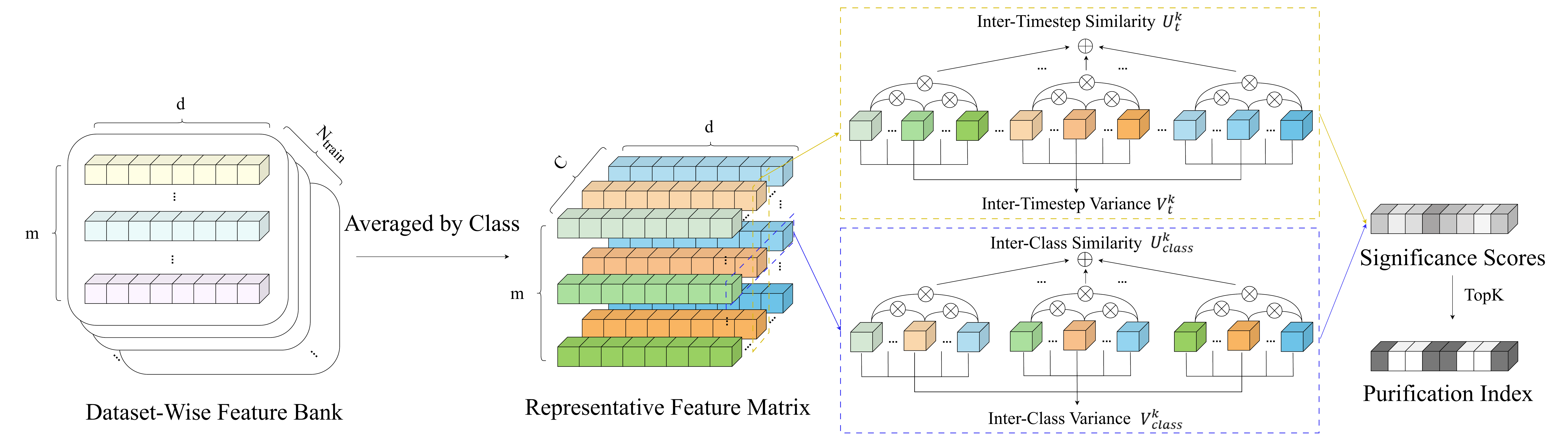}
     \caption{Purification index generation of our proposed class \& timestep-oriented feature purification module.}
    \label{fig:method_puri}
\end{figure*}
\subsection{Multi-Timestep Multi-Stage Diffusion Feature Extraction}
After diffusion feature learning on unlabeled HSI, there exist abundant multi-timestep multi-stage diffusion features that contain both contextual semantics and textual information hidden in the denoising U-Net with different timestep $t$.
% , with a rich hierarchy of information that contains both contextual semantics and textual information, can be extracted from the denoising U-Net with different timestep $t$. 
To model the complex spectral-spatial relations in HSIs, we extract multi-timestep multi-stage features from all stages of the fixed pretrained denoising U-Net decoder with all timesteps, and construct the timestep-wise center and global feature bank by center extraction and average pooling.
% , which consist of multi-stage diffusion features with different timestep $t$.
%Center features are utilized for predicting the labels, and global features assist the corresponding center features in modeling spatial relations with more spatial information from neighboring pixels. 

Specifically, given a labeled patch $x_0^{'} \in \mathcal{R}^{H \times H \times D}$ preprocessed by PCA to $D$ channels, $x_0^{'}$ is corrupted by adding Gaussian noise according to Eq.~\ref{eq2} and obtain $\{x^{'}_{t_i}\}_{i=1}^m$ at a set of timesteps $\{t_i\}_{i=1}^m$ that are sampled from $[0, T]$ at equal intervals. Then, the noisy patches $\{x^{'}_{t_i}\}_{i=1}^m$ are fed into the pretrained denoising U-Net to extract multi-timestep multi-stage diffusion features from all stages of the U-Net decoder.
The different layer features are jointly upsampled to $H \times H$ and then concatenated to get the multi-stage feature $f_{t_i} \in \mathcal{R}^{H \times H \times d}$ at timestep $t_i$. 

For each multi-stage feature $f_{t_i}$, we only reserve the center feature $c_{t_i} \in \mathcal{R}^{1 \times 1 \times d}$ located as $(\frac{H}{2},\frac{H}{2})$ corresponding to the center pixel, and obtain the global feature $g_{t_i} \in \mathcal{R}^{1 \times 1 \times d}$ through global average pooling, which largely reduce computational cost with fewer memories. Following the above process, we construct the timestep-wise center feature bank $\mathcal{B}_c$ and the timestep-wise global feature bank $\mathcal{B}_g$:
\begin{equation}
\mathcal{B}_c = \{ c_{t_i} | i \in \{1,...,m\}, c_{t_i} = center(f_{t_i}) \}
\end{equation}
\begin{equation}
\mathcal{B}_g = \{ g_{t_i} | i \in \{1,...,m\}, g_{t_i} = avgpool(f_{t_i}) \}
\end{equation}

\subsection{Class \& Timestep-oriented Multi-Stage Feature Purification}
Multi-stage features, extracted from the denoising U-Net, embody abundant reconstruction information from the pretraining process of the diffusion model. Despite this richness, the information is not entirely aligned with HSI classification requirements, containing redundant features irrelevant to the task. Furthermore, these features exhibit a degree of repetition among themselves.
%Multi-stage features are directly extracted from the denoising U-Net, containing rich reconstruction information due to the pretraining process of the diffusion model. However, the reconstruction information is not a perfect fit for the classification, which still contains a lot of redundant information that is not relevant to the classification task. Furthermore, there is also some repetitive information among the features. 
Therefore, the class \& timestep-oriented multi-stage purification is proposed to explore multi-stage features by selecting significant channels, aiming to remove the redundant information and reduce the computational cost. 

Before the multi-stage feature purification, we first generate the purification index using the prior of dataset-wise multi-timestep multi-stage feature bank, depicted in Fig.~\ref{fig:method_puri}. Specifically, for a $C$-category HSI classification dataset, $S^j$ is the training samples of category $j$, $j \in \{1, ..., C\}$. The feature banks of samples in $S^j$ are averaged to get the category representative features of category $j$. Representing all training samples by the category representative features can significantly reduce computational overhead. Gathering all the representative features of each category, a representative feature matrix $M \in \mathcal{R}^{m \times C \times d}$ can be obtained,
\begin{equation}
M_{i,j} = \frac{1}{|S^j|} \sum_{x \in S^j} c_{t_i}(x)
\end{equation}
where $c_{t_i}(x)$ is the feature at $t_i$ in the center feature bank of sample $x$, $|S^j|$ is the size of $S^j$.

%Class-oriented significance score. 
In order to purify the discriminative and effective channels for classification, a class-oriented significance score is proposed to evaluate the significance of each channel from inter-class relations. It aims to filter out channels that are highly homogenized and thus have little impact on classification by minimizing the inter-class similarity and maximizing the inter-class variance. For $k^{th}$ index, the class-oriented significance score $\tau_{class}^k$ is formed as:
\begin{equation}
\tau_{class}^k = -\alpha U_{class}^k +(1-\alpha)V_{class}^k
\end{equation}
where $\alpha \in (0,1)$, $U_{class}^k$ denotes the inter-class similarity at index $k$ obtained by summing the average cosine similarities across classes at all the timesteps, and $V_{class}^k$ denotes the inter-class variance at index $k$ which is the sum of the variances across classes at all the timesteps, formed as follows.
\begin{equation}
U_{class}^k = \frac{1}{m} \frac{1}{c^2} \sum_{i=1}^{m} \sum_{p=1}^{c} \sum_{\substack{q=1 \\ q \ne p}}^{c} m_{i,p,k} \cdot m_{i,q,k}
\end{equation}
\begin{equation}
V_{class}^k = \frac{1}{m} \frac{1}{c} \sum_{i=1}^{m} \sum_{p=1}^{c} (M_{i,p,k}-\frac{1}{c}\sum_{q=1}^{c}M_{i,q,k} )^2
\end{equation}

Similarly, to preserve the diversity of features while reducing repetitive information at the timestep dimension different from the class dimension, the timestep-oriented significance score $\tau_{t}^k$ at index $k$ is designed to be calculated as:
\begin{equation}
\tau_{t}^k = -\beta U_t^k +(1-\beta)V_t^k
\end{equation}
where $\beta \in (0,1)$, $U_{t}^k$ denotes the inter-timestep similarity at index $k$ obtained by summing the average cosine similarities across timesteps at all the classes, and $V_{t}^k$ is the inter-timestep variance at index $k$ which is the sum of the variances across classes at all the timesteps, defined as follows. 
\begin{equation}
U_{t}^k = \frac{1}{c} \frac{1}{m^2} \sum_{i=1}^{c} \sum_{p=1}^{m} \sum_{\substack{q=1 \\ q \ne p}}^{m} M_{p,i,k} \cdot M_{q,i,k}
\end{equation}
\begin{equation}
V_{t}^k = \frac{1}{C} \frac{1}{m} \sum_{i=1}^{c} \sum_{p=1}^{m} (m_{p,i,k}-\frac{1}{m}\sum_{q=1}^{m}m_{q,i,k} )^2
\end{equation}

Finally, the class \& timestep-oriented significance score of each index $k \in \{1,..., d\}$ is obtained as the final measurement.
\begin{equation}
\tau^k = \tau_{class}^k +\tau_{t}^k
\end{equation}
We rank the indexes by the class \& timestep-oriented significance score $\tau^k$ and generate the purification index by reserving the indexes of the top $K$ highest $\tau^k$, which indicates the most inter-class and inter-timestep divergence and discrimination. Then, multi-stage features in the timestep-wise center feature bank $\mathcal{B}_c$ multiply the purification index to obtain the purified timestep-wise center feature bank $\mathcal{B}_c^{'}$.
\begin{figure}[t]
    \centering
    \includegraphics[width=9cm]{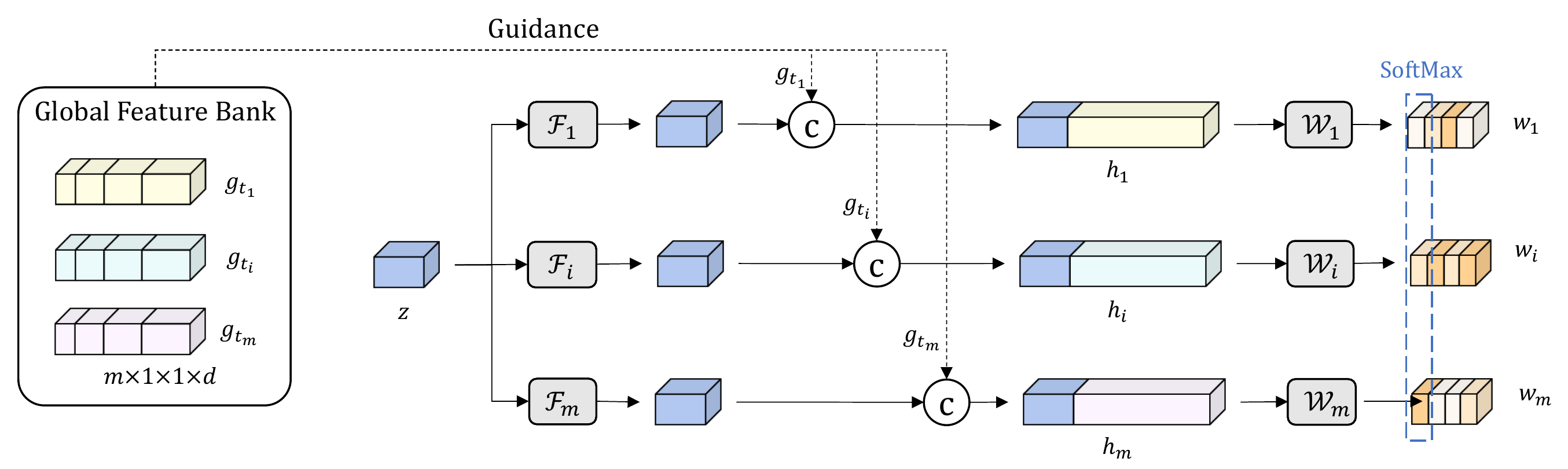}
     \caption{The structure of global feature-guided selective timestep network in our proposed selective timestep feature fusion.}
    \label{fig:method_selective}
\end{figure}
\subsection{Selective Timestep Feature Fusion}
For each labeled HSI patch $x_0^{'}$, multi-timestep features contain textual information and contextual semantics from shallow to deep timesteps. Meanwhile, HSI data have different spectral characteristics for different HSI datasets, resulting in different spectral representation laws of timestep-wise features. Compared to the manual selection of single timestep feature in previous works, we explore the multi-timestep features and perform selective timestep feature fusion to adaptively select different timestep features integrating textual information and contextual semantics and learn comprehensive multi-timestep multi-stage selective representations. The whole process is illustrated in Fig.~\ref{fig:method}.

Specifically, for the purified timestep-wise center feature bank $\mathcal{B}_c^{'} = \{ c_{t_i}^{'} | i \in \{1,...,m\} \}$, our goal is to adaptively select the timestep of features in it for modeling spectral-spatial relations better. 
%We aim to employ a gate mechanism to select proper combinations of timesteps from multi-timestep features with textual information and contextual semantics. Therefore, 
We first fuse all purified center features in $\mathcal{B}_c^{'}$ to obtain the all-timestep feature $c_t \in \mathcal{R}^{1 \times 1 \times K}$ via an element-wise summation:
\begin{equation}
c_t = \sum_{i=1}^m c_{t_i}^{'}
\end{equation}
Then, the all-timestep feature $c_t$ is fed into a simple multi-layer perception (MLP) to obtain a compact feature $z \in \mathcal{R}^{1 \times 1 \times K_r}$ with fewer channels for better efficiency:
\begin{equation}
z = \mathcal{F}_{mlp}(c_t) = W_1(\delta(BN(W_2c_t)))
\end{equation}
where $\delta$ is the ReLU function, $BN$ is the batch normalization, $W_1 \in \mathcal{R}^{K \times K_r}$, $W_2 \in \mathcal{R}^{K_r \times K_r}$, $K_r = K/2$.

However, adaptive timestep selection solely based on timestep-wise center features leads to imprecise choices due to the absence of spatial information from neighboring pixels around the central pixel. Hence, we propose a global feature-guided selective timestep network, designed to incorporate global features for guiding multi-timestep feature selection, thereby enriching spatial information in modeling spatial distributions. And the network is depicted in Fig.~\ref{fig:method_selective}. 
%Namely, global features are sent to the global feature-guided selective timestep network to guide the timestep selection. 
Specifically, given a timestep-wise global feature bank $\mathcal{B}_g = \{ g_{t_i} | i \in \{1,...,m\} \}$, for each purified center feature $c_{t_i}^{'}$, the corresponding global feature $g_{t_i}$ is concatenated with $z$ with the transformation $\mathcal{F}_i$ to obtain $h_i$. Then the linear projection $\mathcal{W}_i$ is applied to the $h_i$ respectively, $i \in \{1,...,m\}$. To adaptively select different timesteps of center features, a softmax operator on the channel-wise digits is used to obtain the selective weights $\{ w_{i} \}_{i=1}^{m}$ guided by the compact feature $z$ and the corresponding global information:
\begin{equation}
w^{'}_i = \mathcal{W}(h_i) = \mathcal{W}([\mathcal{F}_i(z), g_{t_i}])
\end{equation}
\begin{equation}
w^c_i = \frac{e^{w^{'c}_i}}{\sum_{j=1}^m e^{w^{'c}_j}}
\end{equation}
where $\{ \mathcal{F}_{i} \}_{i=1}^{m}$ and $\{ \mathcal{W}_{i} \}_{i=1}^{m}$ are linear projections to align the channel dimension to $K$ channels, and $w^c_i$ is the $c$-th element of the selective weight $w_i$, $c \in \{1,...,K\}$. Finally, the multi-timestep multi-stage selective representation $r_s$ is selected and fused through the selective weights $\{ w_{i} \}_{i=1}^{m}$ on each purified center feature $c_{t_i}^{'}$:
\begin{equation}
r_s = \sum_{i=1}^m w_ic_{t_i}^{'}
\end{equation}

\begin{table}[t]
\centering
\caption{Land-cover types, the number of labeled training samples and testing samples of the Indian Pines dataset.}
\begin{tabular}{c||ccc}
\toprule[1.5pt]
Class&Land Cover Type&Training&Testing\\
\hline \hline  1&Alfalfa&5&41\\
 2&Corn-Notill&143&1285\\
 3&Corn-Mintill&83&747\\
 4&Corn&24&213\\
 5&Grass-Pasture&48&435\\
 6&Grass-Trees&73&657\\
 7&Grass-Pasture-Mowed&3&25\\
 8&Hay-Windrowed&48&430\\
 9&Oats&2&18\\
 10&Soybean-Notill&97&875\\
 11&Soybean-Mintill&245&2210\\
 12&Soybean-Clean&59&534\\
 13&Wheat&20&185\\
 14&Woods&126&1139\\
 15&Buildings-Grass-Trees-Drives&39&347\\
 16&Stone-Steel-Towers&9&84\\
\hline \hline &Total&1024&9225\\
\bottomrule[1.5pt]
\end{tabular}
\label{Table:Indian}
\end{table}

\begin{table}[t]
\centering
\caption{Land-cover types, the number of labeled training samples and testing samples of the PaviaU dataset.}
\begin{tabular}{c||ccc}
\toprule[1.5pt]
Class&Land Cover Type&Training&Testing\\
\hline \hline 1&Asphalt&332&6299\\
 2&Meadows&932&17717\\
 3&Gravel&105&1994\\
 4&Trees&153&2911\\
 5&Painted Metal Sheets&67&1278\\
 6&Bare Soil&251&4778\\
 7&Bitumen&67&1263\\
 8&Self-Blocking Bricks&184&3498\\
 9&Shadows&47&900\\
\hline \hline &Total&2138&40638\\
\bottomrule[1.5pt]
\end{tabular}
\label{Table:Pavia}
\end{table}

%\subsubsection{Lightweight Ensemble Classifiers}
After obtaining the multi-timestep multi-stage selective representation, a lightweight network is needed to predict the classification label. Inspired by \cite{Zhang2021DatasetGANEL}, we train an ensemble of lightweight linear classifiers that takes the dynamic pixel representations as inputs and predicts the classification label of each pixel. Specifically, each classifier is trained independently, consisting of two hidden layers with ReLU activation and batch normalization. When testing a sample, the final predicted label is obtained by majority voting of the ensemble of pixel classifiers, as illustrated in Fig. \ref{fig:method}. This method brings more stability of prediction with a very small cost since the parameters of each classifier are very limited.

\section{Experiments and results}
\review{In this section, we first describe four well-known HSI datasets, including the Indian Pines dataset, the Pavia University dataset, the Houston 2018 dataset, and the WHU-Hi-Longkou dataset. The experimental setting is then introduced including evaluation metrics, a brief introduction of compared state-of-art methods, and implementation details. Then, we conduct quantitative experiments and ablation analysis to evaluate our proposed method.}
\subsection{Datasets Description}
\subsubsection{Indian Pines}
The Indian Pines dataset was acquired in 1992 over an area of Indian pines in North-Western Indiana by Airborne Visible Infrared Imaging Spectrometer (AVIRIS) sensor. It consists of $145 \times 145$ pixels with a spatial resolution of 20 m and 220 spectral bands in the wavelength range of 400 to 2500 nm. There are 200 bands retained for classification (1-103, 109-149, 164-219) after removing the bands affected by noise. The dataset contains 10249 labeled pixels with 16 categories. We use 10\% of the labeled samples for training and the rest for testing. The class name and the number of training and testing samples are listed in Table \ref{Table:Indian}.  

\subsubsection{Pavia University}
The Pavia University (PaviaU) dataset was collected in 2003 by the reflective optics system imaging spectrometer (ROSIS-3) sensor over a part of the city of Pavia, Italy. The dataset consists of $610 \times 340$ pixels with a spatial resolution of 1.3 m and 115 spectral bands in the wavelength range of 430 to 860 nm. 103 out of 115 bands are used for classification after removing 12 noisy bands. The image contains a large number of background pixels, and only 42776 labeled pixels are divided into 9 classes, including asphalt, meadows, gravel, and so on. We use 5\% of the labeled samples for training and the rest for testing. The class name and the number of training and testing samples are listed in Table \ref{Table:Pavia}.

\subsubsection{Houston 2018}
The Houston 2018 dataset, identified as the 2018 IEEE GRSS DFC dataset, was gathered in 2018 by the National Center for Airborne Laser Mapping (NCALM) over the University of Houston campus and its neighboring urban area, including HSI, multispectral LiDAR, and very high-resolution RGB images. The HSI dataset consists of $601 \times 2384$ pixels with a spatial resolution of 1 m and 48 spectral bands in the wavelength range of 380 to 1050 nm. It contains 504856 labeled pixels and 20 classes of interest. We use 5\% of the labeled samples for training and the rest for testing. The class name and the number of training and testing samples are listed in Table \ref{Table:H2018}.
\review{\subsubsection{WHU-Hi-Longkou}
The WHU-Hi-Longkou dataset was acquired in 2018 by an 8-mm focal length Headwall Nano-Hyperspec imaging sensor equipped on a DJ-innovations Matrice 600 Pro UAV platform. It consists of $550 \times 400$ pixels with a spatial resolution of 0.463 m and 270 spectral bands in the wavelength range of 400 to 1000 nm. It contains 204542 labeled samples and 9 object classes. We use 0.5\% of the labeled samples for training and the rest for testing. The class name and the number of training and testing samples are listed in Table \ref{Table:longkou}.}
\begin{table}[t]
\centering
\caption{Land-cover types, the number of labeled training samples and testing samples of the Houston 2018 dataset.}
\begin{tabular}{c||ccc}
\toprule[1.5pt]
Class&Land Cover Type&Training&Testing\\
\hline \hline 1&Healthy Grass&490&9309\\
 2&Stressed Grass&1625&30877\\
 3&Artificial turf&34&650\\
 4&Evergreen trees&680&12915\\
 5&Deciduous trees&251&4770\\
 6&Bare earth&226&4290\\
 7&Water&13&253\\
 8&Residential buildings&1989&37783\\
 9&Non-residential buildings&11187&212565\\
 10&Roads&2293&43573\\
 11&Sidewalks&1702&32327\\
 12&Crosswalks&76&1442\\
 13&Major thoroughfares&2317&44031\\
 14&Highways&493&9372\\
 15&Railways&347&6590\\
 16&Paved parking lots&575&10925\\
 17&Unpaved parking lots&7&139\\
 18&Cars&327&6220\\
 19&Trains&269&5100\\
 20&Stadium seats&341&6483\\
\hline \hline &Total&25242&479614\\
\bottomrule[1.5pt]
\end{tabular}
\label{Table:H2018}
\end{table}

\begin{table}[t]
\centering
\caption{\review{Land-cover types, the number of labeled training samples and testing samples of the WHU-Hi-Longkou dataset.}}
\begin{tabular}{c||ccc}
\toprule[1.5pt]
Class&Land Cover Type&Training&Testing\\
\hline \hline 1     & Corn                & 172      & 34339   \\
2     & Cotton              & 42       & 8332    \\
3     & Sesame              & 15       & 3016    \\
4     & Broad-leaf soybean  & 316      & 62896   \\
5     & Narrow-leaf soybean & 21       & 4130    \\
6     & Rice                & 59       & 11795   \\
7     & Water               & 335      & 66721   \\
8     & Roads and houses    & 36       & 7088    \\
9     & Mixed weed          & 26       & 5203    \\
\hline \hline & Total               & 1022     & 203520 \\
\bottomrule[1.5pt]
\end{tabular}
\label{Table:longkou}
\end{table}
\begin{table*}[ht]
\centering
\caption{Quantitative performance of different classification methods in terms of OA, AA, and $\kappa$ as well as the accuracies for each class on the Indian Pines dataset. The best results are shown in bold.}
%\resizebox{1\textwidth}{!}{
\begin{tabular}{c||ccccccccccc}
\toprule[1.5pt] 
Class & 2-D CNN & 3-D CNN & SSRN & SF & SSFTT & GAHT & 3DCAE & 3DAES & UMSDFL & SpectralDiff & MTMSD \\
\hline \hline
1 & 65.85 & 58.54 & 94.14 & 63.00 & 95.12 & 97.56 & 72.97 & \bf100.00 & 99.10 & \bf100.00 & \bf100.00\\
2 & \bf99.77 & 76.19 & 97.84 & 92.35 & 97.67 & 98.05 & 88.50 & 89.34 & 96.10 & 97.90 & 99.52\\
3 & 81.66 & 77.64 & 97.54 & 86.86 & 98.87 & 98.66 & 87.20 & 95.98 & 95.39 & 98.93 & \bf99.01\\
4 & 96.71 & 52.11 & 90.70 & 88.96 & 91.55 & 95.31 & 84.90 & 95.31 & 97.24 & \bf100.00 & 99.62\\
5 & 85.75 & 93.56 & 97.75 & 92.49 & 96.32 & 95.17 & 90.28 & 88.74 & 94.12 & 94.02 & \bf98.62\\
6 & 97.87 & 98.17 & 99.24 & 99.12 & 99.54 & 99.85 & 97.97 & 99.09 & 99.25 & 99.54 & \bf99.97\\
7 & \bf100.00 & 36.00 & 81.60 & 52.50 & \bf100.00 & \bf100.00 & 56.52 & 56.00 & 88.46 & \bf100.00 & 99.20\\
8 & \bf100.00 & 98.60 & \bf100.00 & 99.16 & \bf100.00 & \bf100.00 & 99.48 & 99.77 & \bf100.00 & \bf100.00 & \bf100.00\\
9 & 50.00 & 55.56 & 74.44 & 41.18 & 88.89 & \bf100.00 & 87.50 & \bf100.00 & 94.44 & \bf100.00 & \bf100.00\\
10 & 35.54 & 82.86 & 94.77 & 93.16 & 97.71 & 94.29 & 86.80 & 87.77 & 95.84 & 98.51 & \bf98.79\\
11 & 88.01 & 90.45 & 98.87 & 92.27 & 98.69 & 99.37 & 96.68 & 96.88 & 99.29 & \bf99.77 & 99.67\\
12 & 98.13 & 62.55 & 97.83 & 85.44 & 98.13 & 96.63 & 80.83 & 83.71 & 93.37 & 91.76 & \bf98.84\\
13 & 99.46 & 88.65 & 99.24 & 99.02 & 97.28 & \bf100.00 & \bf100.00 & \bf100.00 & \bf100.00 & 99.46 & 99.78\\
14 & 99.91 & 99.39 & 99.18 & 96.73 & 99.91 & 97.89 & 99.90 & 99.30 & 95.25 & \bf99.91 & 99.86\\
15 & 91.35 & 86.17 & 93.95 & 83.41 & 98.84 & 97.12 & 96.80 & 98.56 & 99.15 & 98.27 & \bf99.42\\
16 & 86.90 & 45.24 & 98.33 & 93.50 & 95.54 & 94.05 & 84.00 & 97.62 & \bf100.00 & 98.81 & 98.10\\
\hline \hline
OA (\%) & 87.77 & 85.42 & 97.75 & 92.31 & 97.47 & 97.95 & 92.69 & 94.34 & 97.02 & 98.54 & \bf 99.45\\
AA (\%) & 86.06 & 75.10 & 94.71 & 84.95 & 96.57 & 97.75 & 88.15 & 93.00 & 96.00 & 98.56 & \bf 99.40\\
$\kappa$ & 0.8603 & 0.8324 & 0.9743 & 0.9124 & 0.9711 & 0.9766 & 0.9162 & 0.9353 & 0.9660 & 0.9833 & \bf 0.9937\\
\bottomrule[1.5pt]
\end{tabular}
\label{tab:IP}
\end{table*}
\begin{figure*} [thp]
	\centering
	\includegraphics[width=\textwidth]{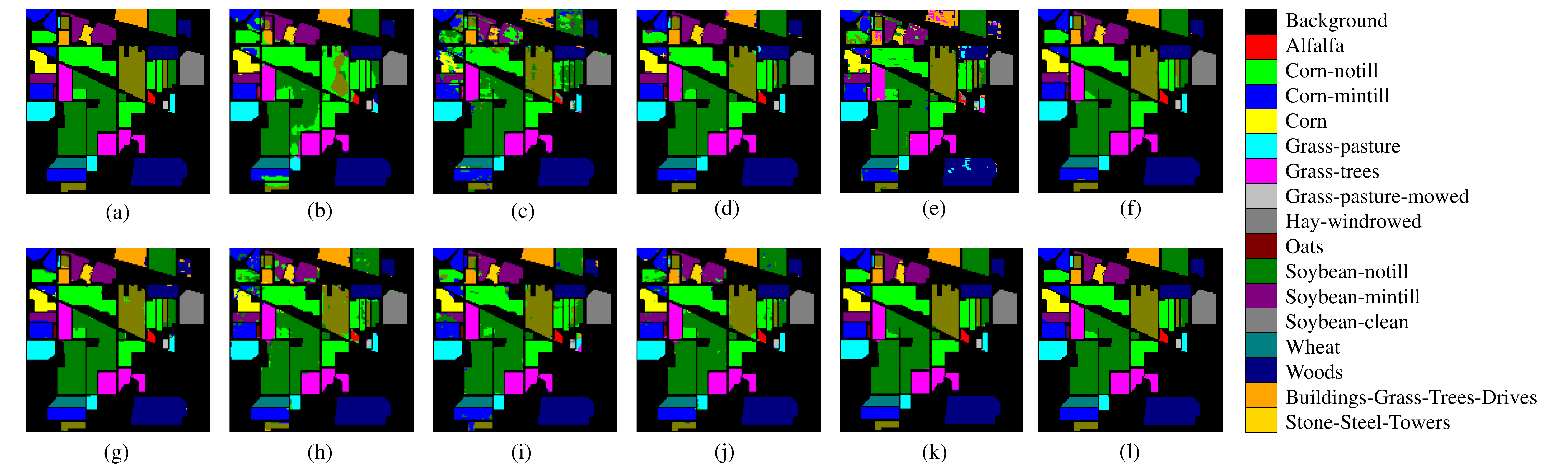}
	\caption{Classification maps obtained by different methods on the Indian Pines dataset. (a) Ground truth. (b) 2-D CNN (OA=87.77\%). (c) 3-D CNN (OA=85.42\%). (d) SSRN (OA=97.75\%). (e) SF (OA=92.31\%). (f) SSFTT (OA=97.47\%). (g) GAHT (OA=97.95\%). (h) 3DCAE (OA=92.69\%). (i) 3DAES (OA=94.34\%). (j) UMSDFL (OA=97.02\%). (k) SpectralDiff (OA=98.54\%). (l) MTMSD (OA=99.45\%).}
	\label{IP_results} 
\end{figure*}
\begin{table*}[t]
\centering
\caption{Quantitative performance of different classification methods in terms of OA, AA, and $\kappa$, as well as the accuracies for each class on the PaviaU dataset. the best results are shown in bold.}
%\resizebox{1\textwidth}{!}{
\begin{tabular}{c||ccccccccccc}
\toprule[1.5pt] 
Class & 2-D CNN & 3-D CNN & SSRN & SF & SSFTT & GAHT & 3DCAE & 3DAES & UMSDFL & SpectralDiff & MTMSD \\
\hline \hline
1 & 99.68 & 97.02 & 98.81 & 96.21 & 99.33 & 99.38 & 94.20 & 97.94 & 99.62 & 99.89 & \bf100.00\\
2 & 99.41 & 99.97 & 99.83 & 99.64 & 99.92 & 99.80 & 99.58 & 99.00 & 99.98 & \bf100.00 & \bf100.00\\
3 & 86.56 & 92.98 & 92.45 & 87.65 & 98.29 & 98.35 & 78.83 & 89.57 & 91.02 & 98.75 & \bf100.00\\
4 & 98.18 & 97.53 & 98.32 & 96.64 & 98.49 & 99.52 & 97.53 & 98.63 & 98.40 & 97.35 & \bf99.59\\
5 & 99.84 & 99.06 & 99.65 & 99.97 & 99.53 & \bf100.00 & \bf100.00 & \bf100.00 & \bf100.00 & 95.93 & \bf100.00\\
6 & \bf100.00 & 99.10 & 99.43 & 99.56 & \bf100.00 & 99.75 & 95.30 & 97.11 & 98.32 & \bf100.00 & \bf100.00\\
7 & 99.84 & 79.10 & 99.76 & 90.30 & 99.13 & 99.60 & 97.42 & 94.77 & 99.61 & \bf100.00 & \bf100.00\\
8 & \bf100.00 & 97.34 & 99.32 & 94.60 & 98.05 & 98.63 & 96.87 & 98.20 & 98.36 & 99.77 & 99.87\\
9 & 98.22 & 95.22 & 99.82 & 98.49 & 95.44 & 99.33 & 98.61 & 97.44 & \bf99.89 & 94.44 & 99.79\\
\hline \hline
OA (\%) & 98.86 & 97.88 & 99.10 & 97.54 & 99.21 & 99.53 & 96.77 & 97.92 & 99.02 & 99.46 & \bf 99.95\\
AA (\%) & 97.97 & 95.26 & 98.60 & 95.88 & 98.69 & 99.37 & 95.37 & 96.96 & 98.36 & 98.46 & \bf 99.92\\
$\kappa$ & 0.9848 & 0.9719 & 0.9881 & 0.9674 & 0.9915 & 0.9937 & 0.9571 & 0.9725 & 0.9870 & 0.9929 & \bf 0.9994\\
\bottomrule[1.5pt]
\end{tabular}
\label{tab:PU}
\end{table*}

\begin{figure*} [thp]
	\centering
	\includegraphics[width=0.95\textwidth]{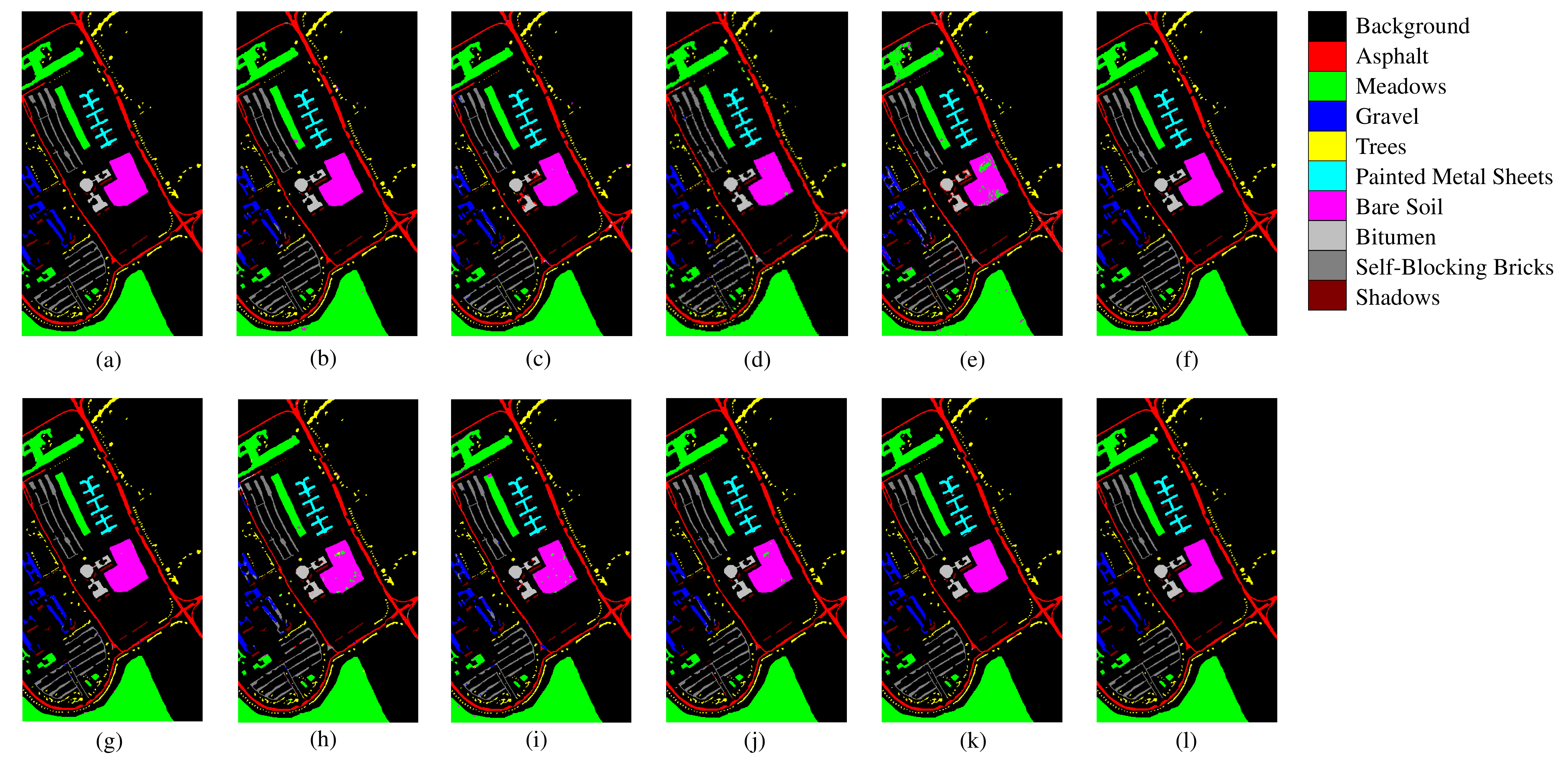}
	\caption{Classification maps obtained by different methods on the PaviaU dataset. (a) Ground truth. (b) 2-D CNN (OA=98.86\%). (c) 3-D CNN (OA=97.88\%). (d) SSRN (OA=99.10\%). (e) SF (OA=97.54\%). (f) SSFTT (OA=99.21\%). (g) GAHT (OA=99.53\%). (h) 3DCAE (OA=96.77\%). (i) 3DAES (OA=97.92\%). (j) UMSDFL (OA=99.02\%). (k) SpectralDiff (OA=99.46\%). (l) MTMSD (OA=99.95\%).}
	\label{PU_results} 
\end{figure*}

\subsection{Experimental Setting}
\subsubsection{Evaluation Metrics}
We evaluate the performance of all methods by three widely used indexes: overall accuracy (OA), average accuracy (AA), and Kappa coefficient ($\kappa$). 
\subsubsection{Comparison with State-of-the-art Methods}
To demonstrate the effectiveness of our proposed method, we compare our classification performance with several state-of-the-art approaches using the most effective setting for these methods.
\begin{itemize}
\item The 2-D CNN\cite{yang2018cnn} architecture contains three 2-D convolution blocks and a softmax layer. Each convolution block consists of a 2-D convolution layer, a BN layer, an avg-pooling layer, and a ReLU activation function.
\item The 3-D CNN\cite{yang2018cnn} contains three 3-D convolution blocks and a softmax layer. Each 3-D convolution block consists of a 3-D convolution layer, a BN layer, a ReLU activation function, and a 3-D convolution layer with step size 2.
\item The SSRN\cite{zhong2018ssrn} is a spectral-spatial residual network based on 3-D CNN and residual connection. Spatial residual blocks and spatial residual blocks are designed to extract discriminative features from HSI data. %The patch size is set to be $7 \times 7$ and the epoch is set to be 200 for all datasets.
\item For SF\cite{hong2021spectralformer}, group-wise spectral embedding and cross-layer adaptive fusion modules in the transformer framework are adopted to capture local spectral representations from neighboring bands.
\item The SSFTT\cite{sun2022ssftt} systematically combines CNN network and transformer structure to exploit spectral-spatial information in the HSI, with a Gaussian weighted feature tokenizer module making the samples more separable.
%\item The HIT\cite{9766028} applies ViTs with 3-D convolutional operation in the HSI data, including an SACP module with two spectral adaptive 3-D convolution kernel and a Conv-Permutater module with ConvPermute and channel-MLP.
\item The GAHT\cite{mei2022gaht} is a end-to-end group-aware transformer method with three-stage hierarchical framework. %The patch size is set to $7 \times 7$ with 300 epochs for all datasets.
\item The 3DCAE\cite{mei2019unsupervised3dcae} is an unsupervised method using an encoder-decoder backbone with 3D convolution operation to learn spectral-spatial features.
\item The 3DAES\cite{3daes} is a semi-supervised method using an autoencoder to extract spectral-spatial features from unlabeled samples and then optimizes the siamese network and classifier using constructed sample pairs.
\item The UMSDFL\cite{zhang2022unsupervisedumsdfl} is an unsupervised method using encoder and decoder with convolutional layers to learn spectral-spatial features. A clustering branch and a multi-layer fusion module are designed to enhance the features.
\item The SpectralDiff\cite{spectraldiff} is a diffusion-based unsupervised method that learns diffusion features through the spectral-spatial diffusion module and feeds diffusion features into the attention-based classification module.
\end{itemize}
\subsubsection{Implementation Details}
The proposed MTMSD was implemented using the Pytorch framework. The patch size is set to $48\times48$, and the dimension of PCA is set to $8/N$, $N$ is the number of spectral bands of the dataset. In the diffusion-pretraining procedure, we use Kullback-Leibler Divergence Loss as the loss function. And the Adam optimizer is adopted with a batch size of 128 and a learning rate of 1e-4, training a total of 40k steps. In the feature-exploring stage, the cross-entropy loss is used in lightweight classifiers. $m$ and $K$ are set to be 19 and 5, respectively. We adopt the Adam Optimizer and the Cosine Annealing as our training schedule. The original learning rate and minimum learning rate are set to be 1e-4 and 5e-6, respectively. The number of epochs is set to 100 for all datasets. We calculate the results fairly by averaging the results of ten repeated experiments with different training sample selections.

\subsection{Quantitative Results and Analysis}
\subsubsection{Classification Results Compared with SOTA Methods}
\begin{figure*} [thp]
	\centering
	\includegraphics[width=\textwidth]{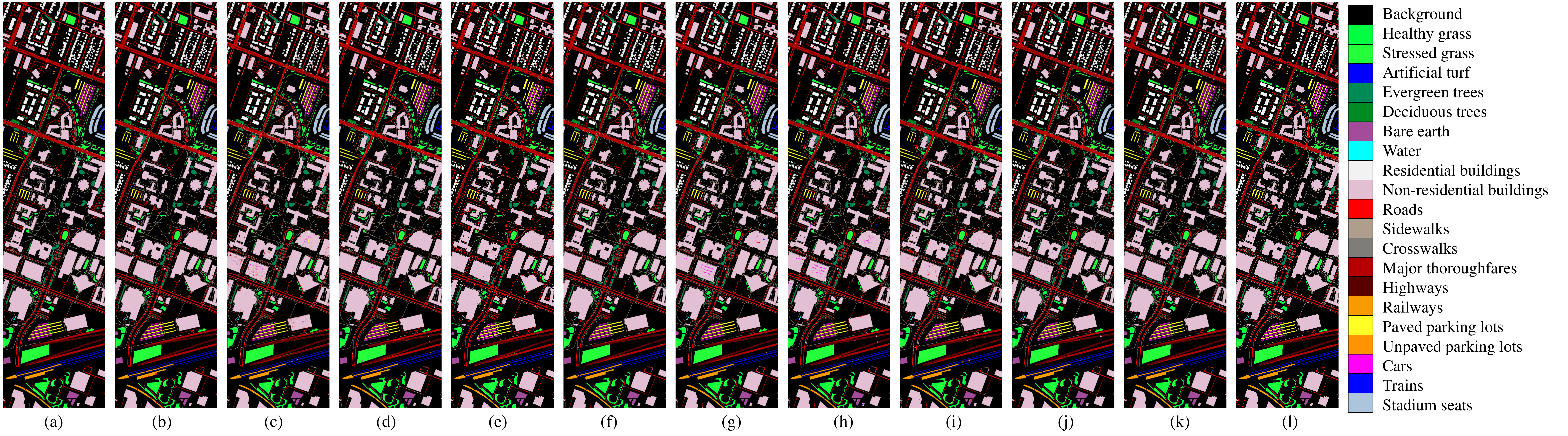}
	\caption{Classification maps obtained by different methods on the Houston 2018 dataset. (a) Ground truth. (b) 2-D CNN (OA=94.84\%). (c) 3-D CNN (OA=86.88\%). (d) SSRN (OA=91.61\%). (e) SF (OA=90.65\%). (f) SSFTT (OA=95.48\%). (g) GAHT (OA=96.69\%). (h) 3DCAE (OA=90.34\%). (i) 3DAES (OA=90.39\%). (j) UMSDFL (OA=95.38\%). (k) SpectralDiff (OA=95.28\%). (l) MTMSD (OA=98.29\%).}
	\label{HU18_results} 
\end{figure*}

\begin{figure*} [thp]
	\centering
	\includegraphics[width=0.95\textwidth]{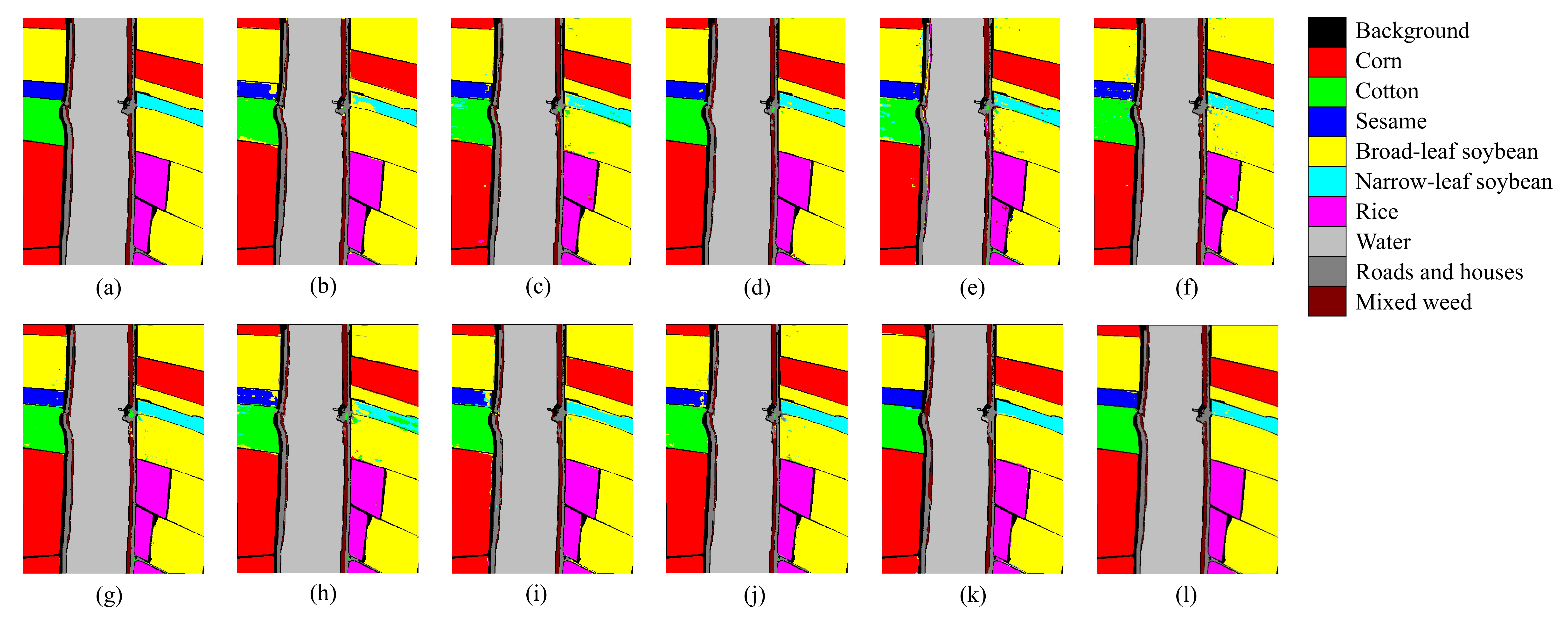}
    \vspace{-1em}
	\caption{\review{Classification maps obtained by different methods on the WHU-Hi-Longkou dataset. (a) Ground truth. (b) 2-D CNN (OA=98.58\%). (c) 3-D CNN (OA=98.48\%). (d) SSRN (OA=99.28\%). (e) SF (OA=97.47\%). (f) SSFTT (OA=99.02\%). (g) GAHT (OA=99.19\%). (h) 3DCAE (OA=97.86\%). (i) 3DAES (OA=98.54\%). (j) UMSDFL (OA=98.99\%). (k) SpectralDiff (OA=98.71\%). (l) MTMSD (OA=99.61\%).}}
	\label{Longkou_results} 

\end{figure*}

\review{Quantitative classification results in terms of class-specific accuracy, OA, AA, and $\kappa$ of the compared methods on the Indian
Pines, PaviaU, Houston 2018 and Longkou datasets are listed in Table \ref{tab:IP}, \ref{tab:PU}, \ref{tab:HH} and \ref{tab:Longkou}, respectively. And the classification maps of all methods are shown in Fig. \ref{IP_results}, \ref{PU_results}, \ref{HU18_results} and \ref{Longkou_results}.}
\begin{table*}[ht]
\centering
\caption{Quantitative performance of different classification methods in terms of OA, AA, and $\kappa$ as well as the accuracies for each class on the Houston 2018 dataset. the best results are shown in bold.}
%\resizebox{0.95\textwidth}{!}{
\begin{tabular}{c||ccccccccccc}
\toprule[1.5pt] 
Class & 2-D CNN & 3-D CNN & SSRN & SF & SSFTT & GAHT & 3DCAE & 3DAES & UMSDFL & SpectralDiff & MTMSD \\
\hline \hline
1  & 87.12 & 82.02 & 86.30 & \bf92.36 & 79.93 & 79.50  & 91.60 & 81.80 & 88.99 & 83.47 & 88.87  \\
2  & 92.05 & 96.64 & 95.32 & 95.08 & 93.44 & 96.55  & 93.92 & 93.30 & \bf97.53 & 92.48 & 96.29  \\
3  & 96.92 & 96.00 & 99.72 & 96.21 & 99.66 & \bf100.00 & 97.60 & 89.23 & 99.23 & 99.38 & 99.93  \\
4  & 98.05 & 95.55 & 97.49 & 98.48 & 96.64 & 97.62  & 95.92 & 94.39 & 97.98 & 97.01 & \bf99.32  \\
5  & 87.25 & 79.16 & 86.09 & 91.87 & 90.11 & 95.01  & 85.79 & 81.95 & 92.51 & 86.25 & \bf97.01  \\
6  & 95.36 & 97.51 & 98.48 & 99.62 & 99.62 & 99.91  & 98.50 & 97.72 & 99.58 & 99.58 & \bf100.00 \\
7  & 96.44 & 71.94 & 93.68 & 27.01 & 85.45 & 95.65  & 61.15 & 83.40 & 96.85 & 87.35 & \bf97.40  \\
8  & 97.15 & 88.62 & 91.62 & 95.42 & 98.72 & 99.18  & 91.15 & 91.27 & 94.20 & 98.73 & \bf99.81  \\
9  & 98.24 & 92.80 & 97.88 & 98.60 & 99.09 & 99.35  & 95.12 & 97.12 & 99.07 & 99.31 & \bf99.72  \\
10 & 93.88 & 71.61 & 81.72 & 88.22 & 91.15 & 92.64  & 76.72 & 76.90 & 88.61 & 87.73 & \bf96.63  \\
11 & 75.85 & 73.17 & 69.44 & 27.01 & 80.97 & 85.73  & 70.22 & 71.14 & 82.28 & 79.33 & \bf93.96  \\
12 & 12.55 & 11.10 & 0.51  & 31.46 & 41.69 & 33.43  & 4.79 & 3.19 & 42.76 & 37.24 & \bf55.30  \\
13 & 85.24 & 69.12 & 84.59 & 91.79 & 94.49 & 96.79  & 90.52 & 84.62 & 93.26 & 95.64 & \bf97.87  \\
14 & 77.12 & 96.18 & 89.65 & 92.91 & 96.97 & 99.17  & 87.23 & 95.41 & 97.02 & 98.09 & \bf99.27  \\
15 & 94.45 & 98.98 & 99.34 & 99.33 & 99.30 & \bf99.92  & 99.06 & 97.31 & 99.59 & 99.47 & \bf99.92  \\
16 & 93.43 & 90.40 & 91.00 & 96.36 & 97.84 & 98.15  & 92.01 & 87.79 & 96.36 & 98.42 & \bf99.84  \\
17 & 64.75 & 20.86 & 0.00  & 22.78 & 69.21 & 90.65  & 0.00 & 46.04 & \bf100.00 & 84.17 & \bf100.00  \\
18 & 91.70 & 89.05 & 93.66 & 91.61 & 93.07 & 97.85  & 90.43 & 83.89 & 94.24 & 93.91 & \bf99.60  \\
19 & 96.88 & 95.45 & 96.92 & 96.53 & 97.97 & 99.84  & 96.09 & 94.55 & 99.59 & 98.47 & \bf99.99  \\
20 & 99.83 & 93.92 & 99.12 & 99.77 & 99.96 & \bf100.00 & 97.27 & 96.30 & 99.89 & \bf100.00 & \bf100.00  \\
\hline \hline
OA (\%) & 93.38 & 86.88 & 91.61 & 90.65 & 95.48 & 96.69  & 90.34 & 90.39 & 95.38 & 95.28 & \bf98.29 \\
AA (\%) & 86.71 & 80.50 & 82.63 & 80.75 & 90.26 & 92.85  & 80.76 & 82.37 & 92.98 & 90.80 & \bf96.04\\
$\kappa$ & 0.9137 & 0.8313 & 0.8906 & 0.8784 & 0.9412 & 0.9570 & 0.8751 & 0.8748 & 0.9398 & 0.9385 & \bf0.9777\\
\bottomrule[1.5pt]
\end{tabular}
\label{tab:HH}
\end{table*}

\begin{table*}[t]
\centering
\caption{\review{Quantitative performance of different classification methods in terms of OA, AA, and $\kappa$, as well as the accuracies for each class on the WHU-Hi-Longkou dataset. the best results are shown in bold.}}
%\resizebox{1\textwidth}{!}{
\begin{tabular}{c||ccccccccccc}
\toprule[1.5pt] 
Class & 2-D CNN & 3-D CNN & SSRN & SF & SSFTT & GAHT & 3DCAE & 3DAES & UMSDFL & SpectralDiff & MTMSD \\
\hline \hline
1       & 99.44          & 99.55  & 99.80  & 99.76  & 99.85  & 99.91  & 99.71  & 98.85  & 99.93  & 99.53           & \textbf{99.97}  \\
2       & 95.44          & 93.18  & 98.82  & 89.20  & 95.96  & 98.99  & 95.07  & 98.78  & 97.40  & 97.49           & \textbf{99.67}  \\
3       & 81.76          & 93.83  & 91.11  & 97.25  & 93.50  & 96.68  & 85.18  & 80.64  & 91.25  & \textbf{100.00} & 97.67           \\
4       & \textbf{99.99} & 98.70  & 99.71  & 98.42  & 99.04  & 99.55  & 98.26  & 99.71  & 99.13  & 99.46           & 99.81           \\
5       & 75.45          & 83.74  & 94.04  & 83.80  & 92.42  & 95.96  & 60.39  & 84.02  & 94.19  & 96.39           & \textbf{97.31}  \\
6       & 98.66          & 98.88  & 99.89  & 97.69  & 99.34  & 99.77  & 99.49  & 98.76  & 99.76  & 99.33           & \textbf{99.86}  \\
7       & 99.96          & 99.99  & 99.97  & 99.98  & 99.99  & 99.99  & 99.99  & 99.94  & 99.99  & 99.85           & \textbf{99.99}  \\
8       & \textbf{98.41} & 96.66  & 96.33  & 89.45  & 97.19  & 96.46  & 96.67  & 95.40  & 97.95  & 86.82           & 96.77           \\
9       & 91.08          & 94.77  & 93.95  & 73.44  & 96.16  & 86.51  & 92.89  & 89.54  & 88.93  & 87.28           & \textbf{96.24}  \\
\hline \hline
OA (\%) & 98.58          & 98.48  & 99.28  & 97.47  & 99.02  & 99.19  & 97.86  & 98.54  & 98.99  & 98.71           & \textbf{99.61}  \\
AA (\%) & 93.36          & 95.37  & 97.07  & 92.11  & 97.05  & 97.09  & 91.96  & 93.96  & 96.20  & 96.24           & \textbf{98.59}  \\
$\kappa$ & 0.9812         & 0.9801 & 0.9905 & 0.9668 & 0.9872 & 0.9893 & 0.9718 & 0.9807 & 0.9868 & 0.9830          & \textbf{0.9949}\\
\bottomrule[1.5pt]
\end{tabular}
\label{tab:Longkou}
\end{table*}

Compared with other methods, our proposed MTMSD achieves the highest OA, AA, and $\kappa$ on \review{four} datasets. According to the results, the CNN-based supervised methods, 2-D CNN, 3-D CNN, and SSRN, obtain good performance owing to their ability to capture local spatial information. Besides, since transformers are capable of capturing sequential information, transformer-based supervised methods, SF, SSFTT, and GAHT, also achieve competitive performance. 
%The GAHT method achieves the best results in CNN-based and transformer-based supervised methods because it balances the local information and the global dependencies. 
%However, the methods that learn explicit mappings have bottlenecks in learning label-agnostic information and complex relations, which limits performance. 
Unsupervised methods, 3DCAE, 3DAES, and UMSDFL are proposed to tackle the problem of limited samples by learning representative features without any labeled samples. Limited by the model architecture, the features learned in an unsupervised manner are not discriminative enough for HSI classification lacking high-level information. Therefore, the performance of 3DCAE and 3DAES is even lower than that of some explicit learning methods. SpetralDiff introduces the diffusion model to HSI classification and achieves competitive performance due to the advantages of diffusion features. However, the feature used in the method is from a single timestep and a single layer with the loss of some key information, which limits the performance. Our proposed MTMSD explores the value of multi-timestep multi-stage diffusion features through class \& timestep-oriented multi-stage feature purification and selective timestep feature fusion, effectively modeling complex spectral-spatial relations due to the adaptive integration of contextual semantics and textual details.
\review{Thus, our MTMSD outperforms all the previous methods on four datasets: Indian Pines, PaviaU, Houston 2018, and Longkou.} Notably, the classification performance of the Houston 2018 Dataset is largely improved compared with the previous SOTA method in terms of OA (98.29\% versus 96.69\%), AA (96.04\% versus 92.98\%), and $\kappa$ (0.9777 versus 0.9570), which especially demonstrate our effectiveness.

\subsubsection{Classification Results with Different Proportions of Training Samples}
\begin{figure*}[thp]
    \centering
    \subfigure[\label{fig:ip_results}]{
		\includegraphics[width=0.23\textwidth]{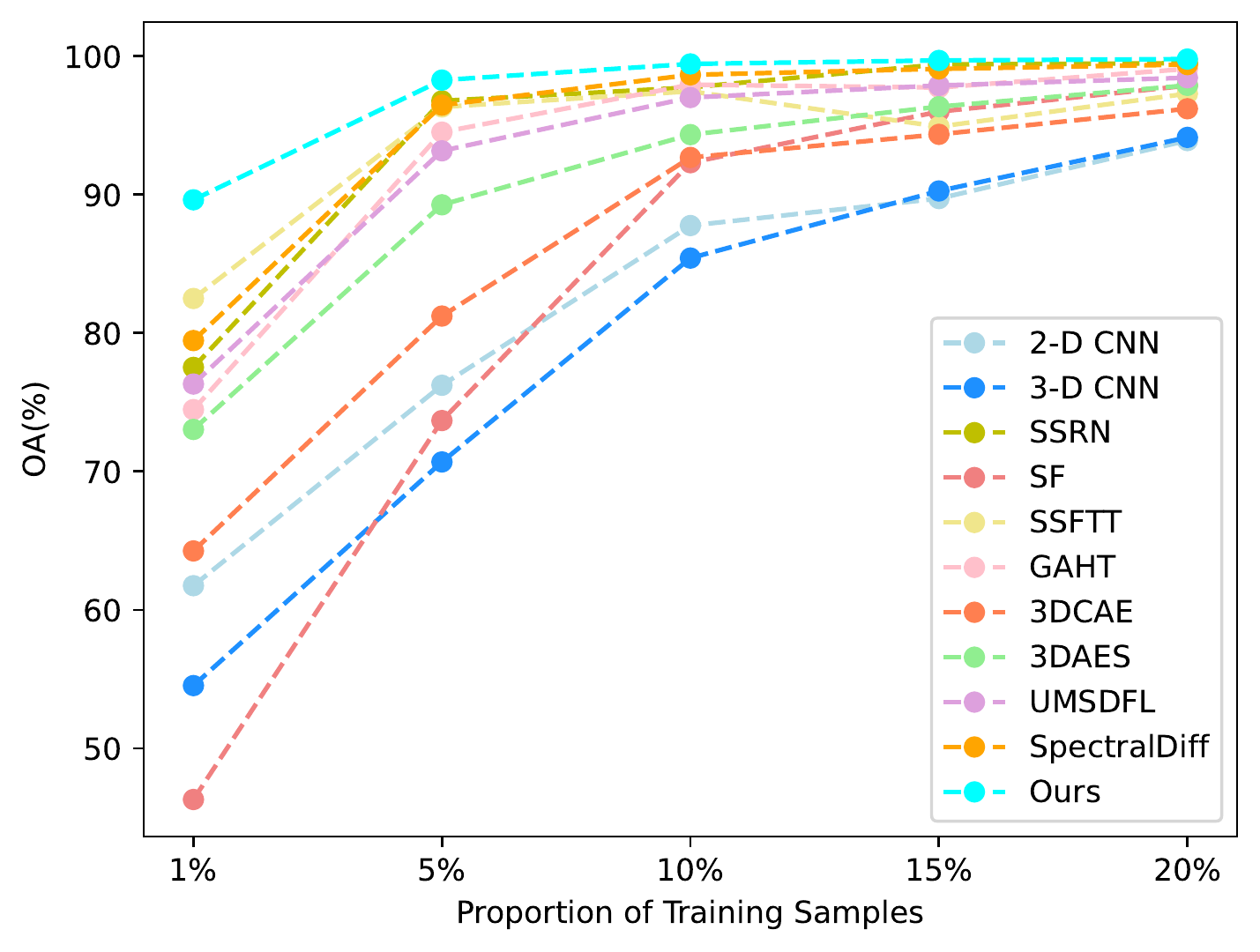}}
    \subfigure[\label{fig:pu_results}]{
		\includegraphics[width=0.23\textwidth]{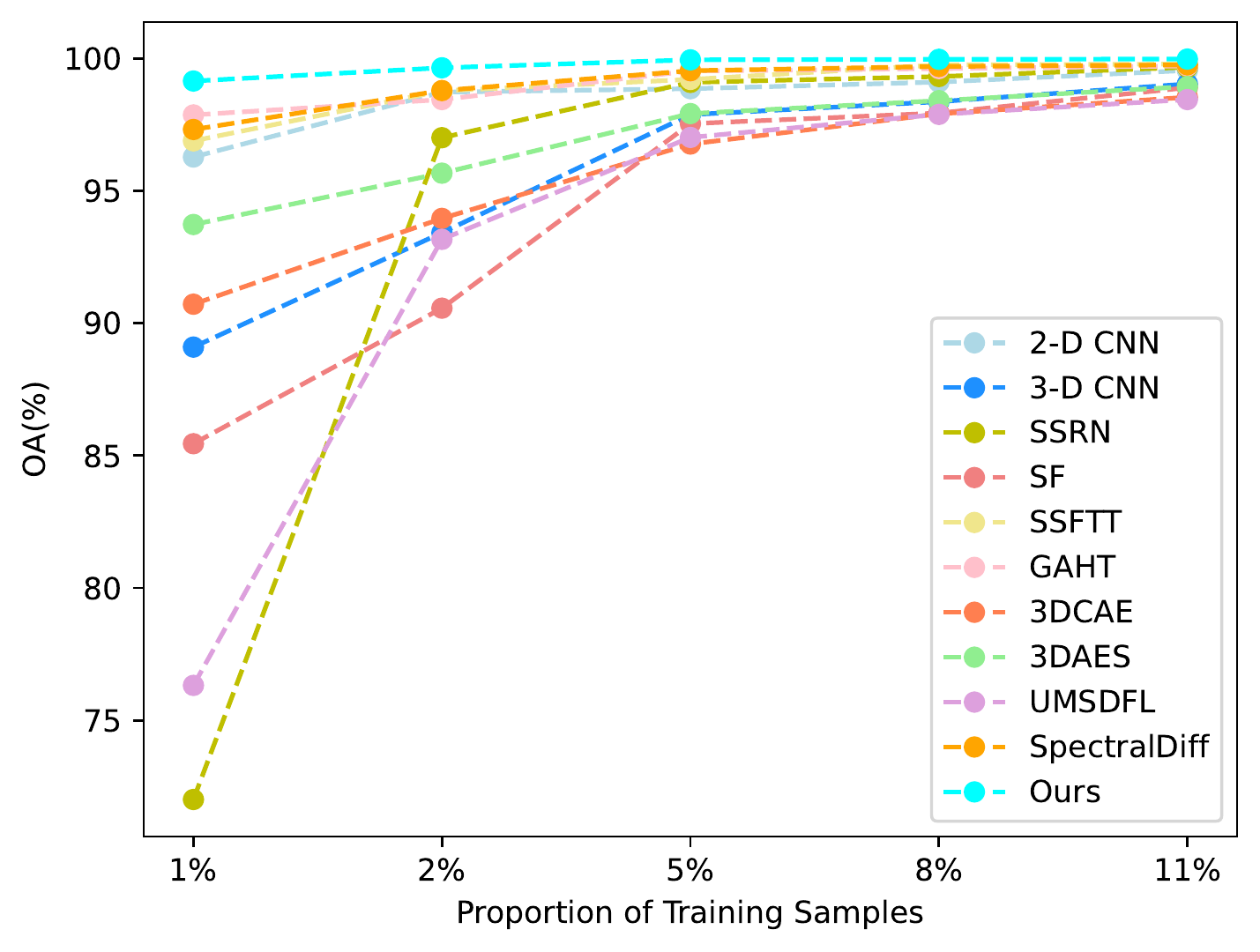}}
  \subfigure[\label{fig:hu18_results}]{
		\includegraphics[width=0.23\textwidth]{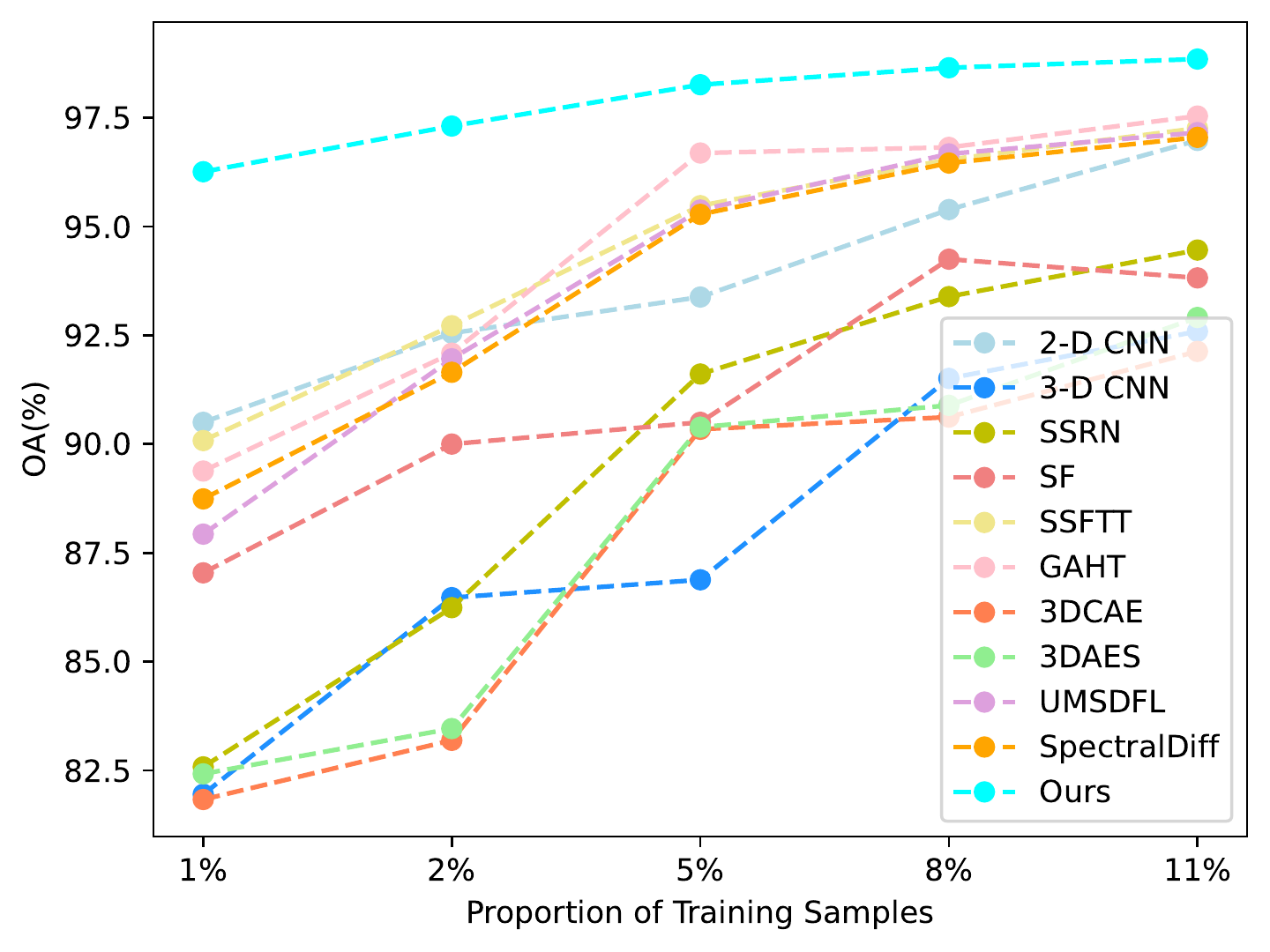}}
  \subfigure[\label{fig:longkou_results}]{
		\includegraphics[width=0.23\textwidth]{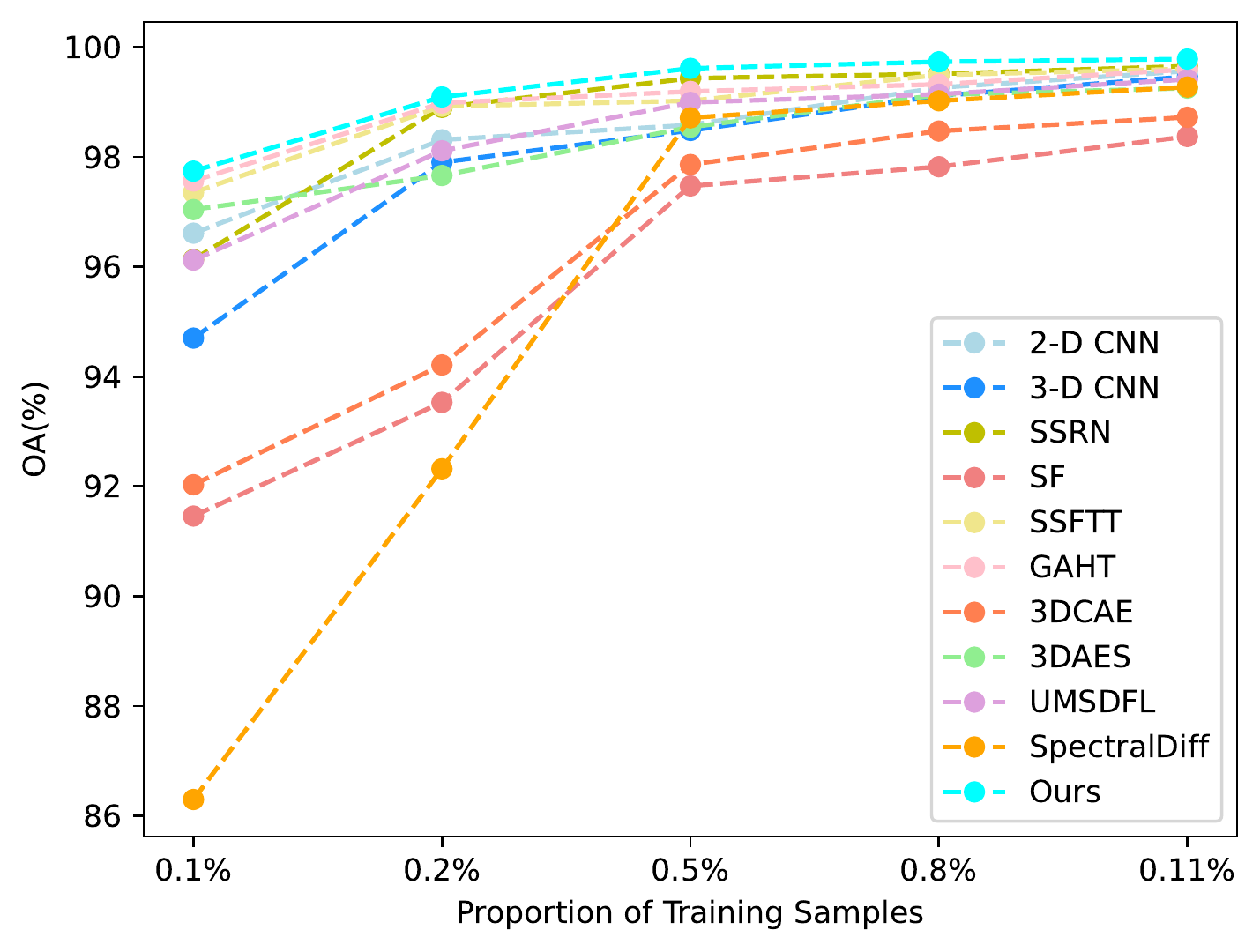}}
    \caption{\review{Classification performance of the compared methods with different proportions of training samples on four datasets. (a) Indian Pines. (b) PaviaU. (c) Houston 2018. (d) Longkou.}}
    \label{fig:results}
\end{figure*}

The classification results with different proportions of training samples are shown in Fig. \ref{fig:results}. \review{It can be observed that the performance increases with the percentages of training samples. Our method outperforms the compared method consistently in terms of OA on four datasets. Especially, using only 2\% of the training sample, our method achieves comparable results to other methods using 5\% of the training sample on the Houston 2018 dataset.}

\subsection{Ablation Studies}
In this section, we analyze the effect of the components in our method. 
\subsubsection{Ablation for Class \& Timestep-oriented Multi-Stage Feature Purification}
\review{Table~\ref{tab:ctmsfp} presents the effect of class \& timestep-oriented multi-stage feature purification (CTMSFP) across the four datasets. The multi-stage features extracted from the diffusion model are not entirely aligned with the HSI classification task. Therefore, the CTMSFP is proposed to reduce the redundancy of the features and maintain efficiency. All experiments are conducted on an RTX 3090 GPU, with a consistent batch size of 64 used during training. As shown in Table~\ref{tab:ctmsfp}, the results demonstrate that CTMSFP significantly reduces the number of parameters and the GPU memory consumption during training. This reduction is due to the channel-wise purification performed by CTMSFP before the features are input into the model, which substantially decreases the size of both the input tensor and the model structure. Furthermore, the average inference time is reduced by 25\%, effectively enhancing the model's computational efficiency, while still ensuring a slight performance improvement. We also compare the performance and efficiency of another diffusion-based HSI classification model, SpectralDiff, which utilizes single-timestep single-stage diffusion features. Our method performs independent linear transformations on features from different timesteps, allowing tailored transformations for diverse multi-timestep feature patterns, resulting in a larger number of parameters. However, our method with CTMSFP significantly leads to performance, inference speed, and GPU memory consumption.}

%Our dynamic feature fusion module contains $n$ sets of dynamic weights which impact the classification performance. We reveal that the best performance is achieved when $n$ is taken as 3. With smaller $n$, only part of the useful information is obtained from the feature bank, which is insufficient for classification. On the other hand, larger $n$ will bring more parameters and excess information.

%\subsubsection{Ablation for Ensemble Classifiers}
%An ensemble of $k$ networks is trained for classification by majority voting in the finetuning stage to reduce the effect of random sampling and make the model more robust. We analyze the effect of the number of networks on classification performance. As shown in Table , the indexes slightly improve while the variance decreases with the increase of $k$. A larger $k$ will bring better stability, but also greater computational cost. When $k=$, the best performance and stability are almost achieved with limited computational cost brought.  

\begin{table*}[t]
\renewcommand{\arraystretch}{1.3}
\centering
\vspace{-1em}
\caption{\review{Ablation for Class \& Timestep-oriented Multi-Stage Feature Purification (CTMSFP) on four datasets in terms of OA, inference time (IT), parameters and GPU memory. The best result are shown in bold.}}
%\resizebox{\columnwidth}{!}{%
\begin{tabular}{l|cccccccc|c|c}
\toprule[1pt] 
\multirow{2}{*}{Method}                                      & \multicolumn{2}{c}{Indian Pines} & \multicolumn{2}{c}{PaviaU}      & \multicolumn{2}{c}{Houston 2018} & \multicolumn{2}{c|}{Longkou}    & \multirow{2}{*}{Param. (M)} & \multirow{2}{*}{GPU memory (G)} \\ \cline{2-9}
                                                             & OA (\%)         & IT (s)         & OA (\%)        & IT (s)         & OA (\%)        & IT (s)          & OA (\%)        & IT (s)         &                             &                                 \\ \hline
SpectralDiff                                                 & 98.54           & 10.92          & 99.46          & 22.41          & 95.28          & 268.88          & 98.71          & 178.25         & \textbf{1.38}               & 3.09                            \\
\begin{tabular}[c]{@{}l@{}}MTMSD (w/o CTMSFP)\end{tabular} & 99.39           & 4.55           & 99.90          & 22.78          & 98.20          & 215.20          & 99.51          & 85.14          & 55.91                       & 1.28                   \\
MTMSD                                                        & \textbf{99.45}  & \textbf{3.52}  & \textbf{99.95} & \textbf{12.41} & \textbf{98.29} & \textbf{168.52} & \textbf{99.61} & \textbf{61.53} & 20.18                       & \textbf{0.46}                   \\ 
\bottomrule[1pt]
\end{tabular}%
%}
\label{tab:ctmsfp}
\end{table*}

\subsubsection{Ablation for Selective Timestep Feature Fusion}
The ablation results for the selective timestep feature fusion are shown in Table. \ref{tab:sf}. In the manual selection method, we extract features from a single timestep for classification. The optimal timestep is selected for each dataset, and the best result is shown in the table. However, using the features from only one single timestep leads to the loss of abundant spectral-spatial information. Although the optimal timestep is chosen for each dataset, it lacks adequate information, only containing textural features or semantics to model spectral-spatial relations, and is not flexible enough to accommodate different patch data, both of which limit performance.
The average fusion method considers features from different timestep $t$, obtaining better results than manual selection. However, assigning a uniform selection weight to features across all the timesteps indiscriminately does not fully harness their potential because the significance of features from different $t$ for different data instances is heterogeneous. 
%The importance of each individual channel is also varying within a single timestep feature. 
Therefore, the proposed selective fusion is more optimal for feature fusion. Compared with the average fusion, our selective timestep feature fusion achieves better performance by assigning the vital timestep features higher selecting weights. Furthermore, compared with no global-feature guidance, our proposed selective timestep fusion with global-feature guidance improves further due to the supplement of global information that enhances the ability to represent spatial distributions.

\begin{table}[t]
    \centering
    \renewcommand{\arraystretch}{1.3}
    \caption{\review{OA (\%) of the proposed MTMSD with different feature fusion and global-feature guidance on four datasets. the best results are shown in bold.}}
    \resizebox{0.49\textwidth}{!}{
    \begin{tabular}{c|c||cccc}
        \toprule[1pt]
         Method & Guidance & Indian Pines & PaviaU & Houston 2018 & Longkou \\
         \hline
         Manual Selection & \XSolid & 98.17 & 99.40 & 94.57 & 99.08\\
         \cline{1-2}
         Average Fusion & \XSolid & 99.03 & 99.78 & 97.85 & 99.39\\
         \cline{1-2}
        \multirow{2}{*}{Selective Fusion} & \XSolid  & 99.31 & 99.90 & 98.07 & 99.53\\
         & \Checkmark & \bf99.45 &  \bf99.95 &  \bf98.29 & \bf99.61\\
        \bottomrule[1pt]
    \end{tabular}
    }
    \label{tab:sf}
\end{table}

\subsection{Discussion and Visualization}
\subsubsection{Analysis of Features from Multiple Timesteps}
\begin{figure}[t]
    
    \centering
    \includegraphics[width=7cm]{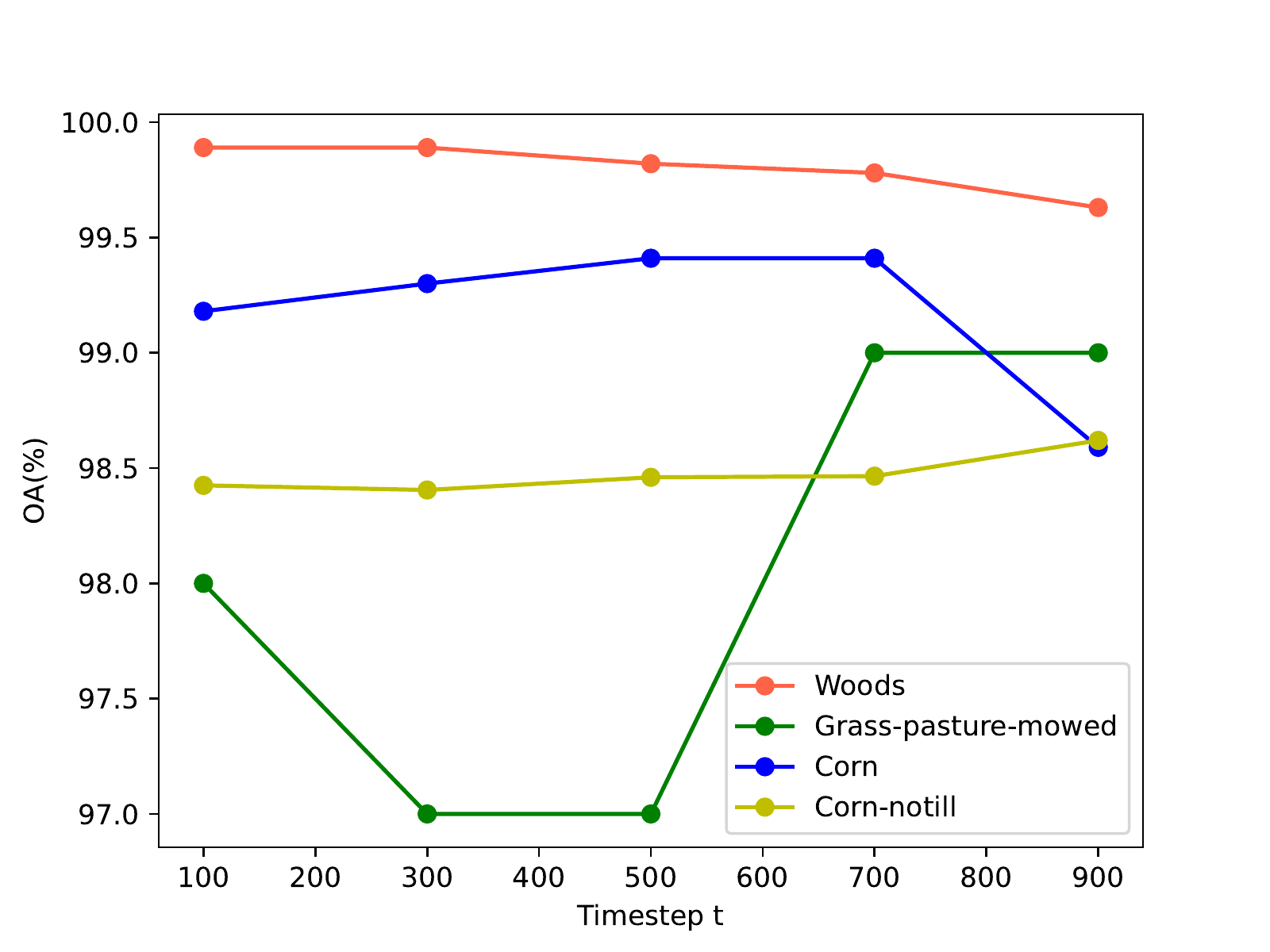}
    \caption{Variation of classification results for four classes with the change of timestep on the Indian Pines dataset.}
    \label{fig:results_with_t}    
\end{figure}

To analyze the features extracted from different timestep $t$, we record the change of the classification performance when changing $t$. For easy understanding, we choose 4 of the 16 classes to show their changes, as illustrated in Fig. \ref{fig:results_with_t}. The performance for each class behaves differently as the $t$ increases. Features of larger $t$ are more sensitive to class "corn-notill" and features of smaller $t$ are more informative to class "woods". For class "Grass-pasture-mowed" and class "corn", features extracted at the intermediate $t$ is the most discriminative. Thus, an appropriate fusion of features at different $t$ is vital for accurate performance.

\begin{figure}[t]
    
    \centering
    \subfigure[\label{fig:retained}]{
		\includegraphics[width=0.48\textwidth]{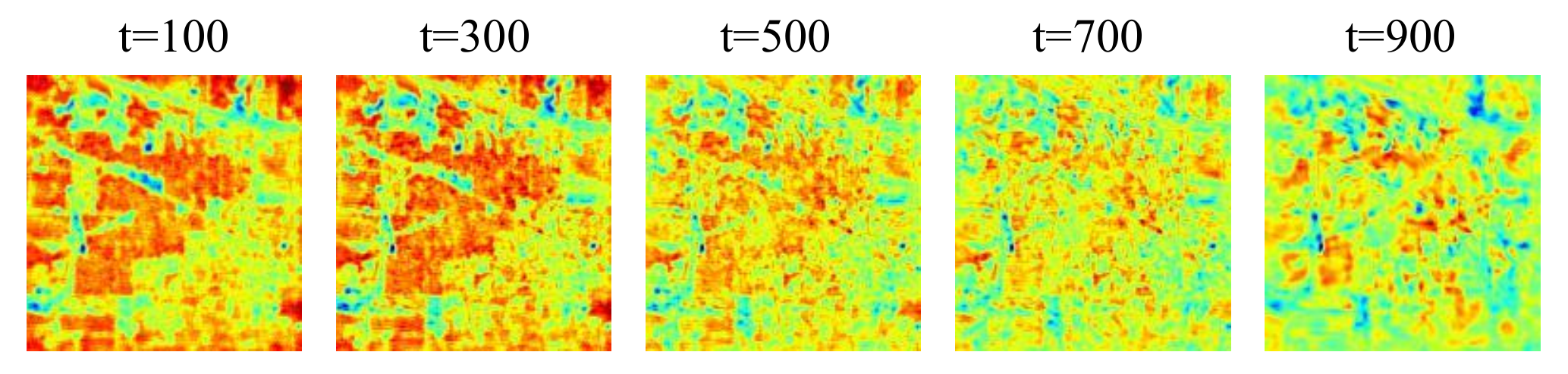}}
    \subfigure[\label{fig:removed}]{
		\includegraphics[width=0.48\textwidth]{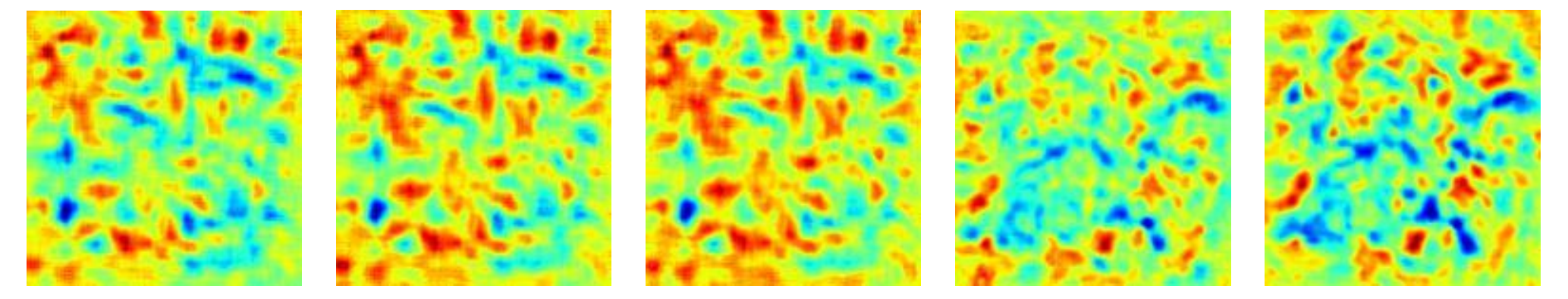}}
    \caption{Feature visualization of retained and removed channels after class \& timestep-oriented multi-stage purification at different timestep $t$. (a) Retained. (b) Removed.}
    \label{fig:visualization_1}
\end{figure}

\begin{figure}[t]   
    \centering
    \subfigure[\label{fig:raw}]{
		\includegraphics[width=0.072\textwidth]{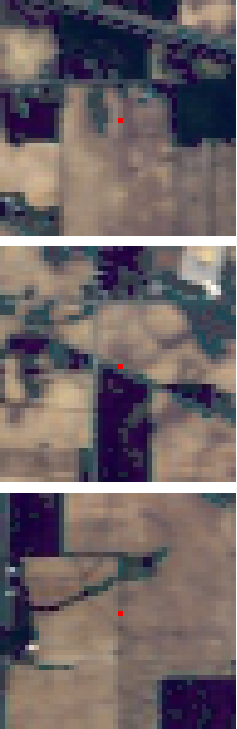}}
    \subfigure[\label{fig:gt}]{
		\includegraphics[width=0.072\textwidth]{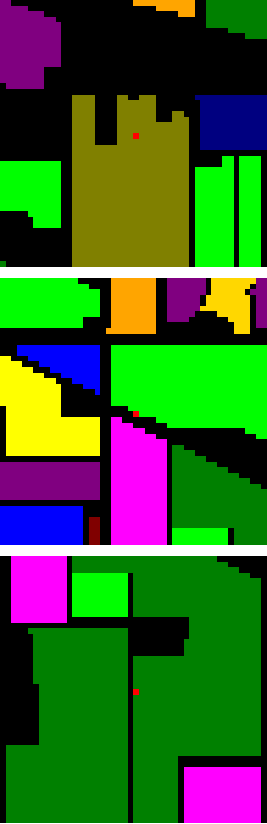}}
    \subfigure[\label{fig:large_w}]{
		\includegraphics[width=0.147\textwidth]{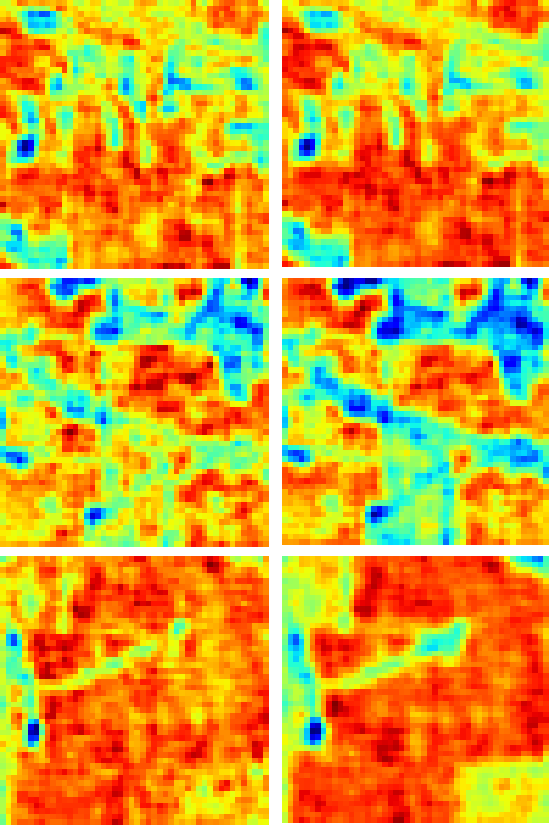}}
    \subfigure[\label{fig:smaller_w}]{
		\includegraphics[width=0.147\textwidth]{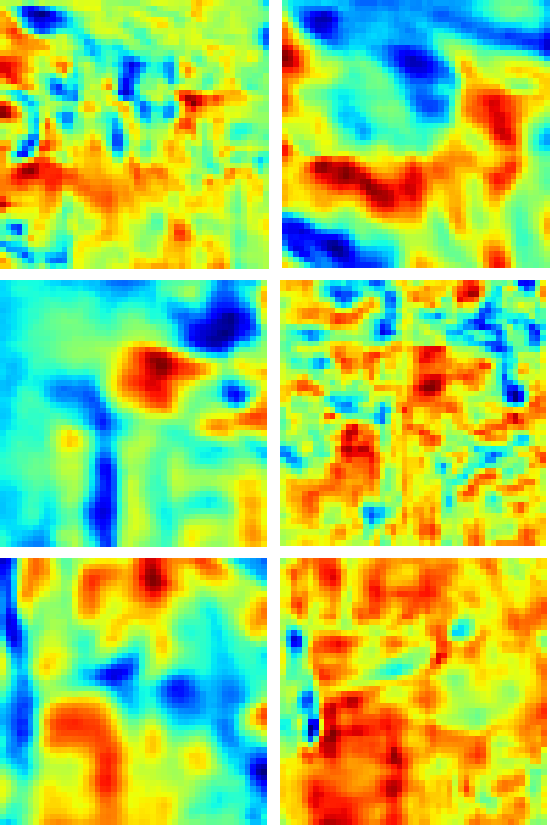}}
    \caption{Feature visualization of different timestep $t$ with higher and lower selection weights. (a) Pseudocolor images. (b) Ground truth. (c) Higher weights. (d) Lower weights. }
    \label{fig:visualization_2}
\end{figure}
\begin{figure*}[t]
    \centering
    \subfigure[\label{fig:results_with_size}]{
		\includegraphics[width=0.32\textwidth]{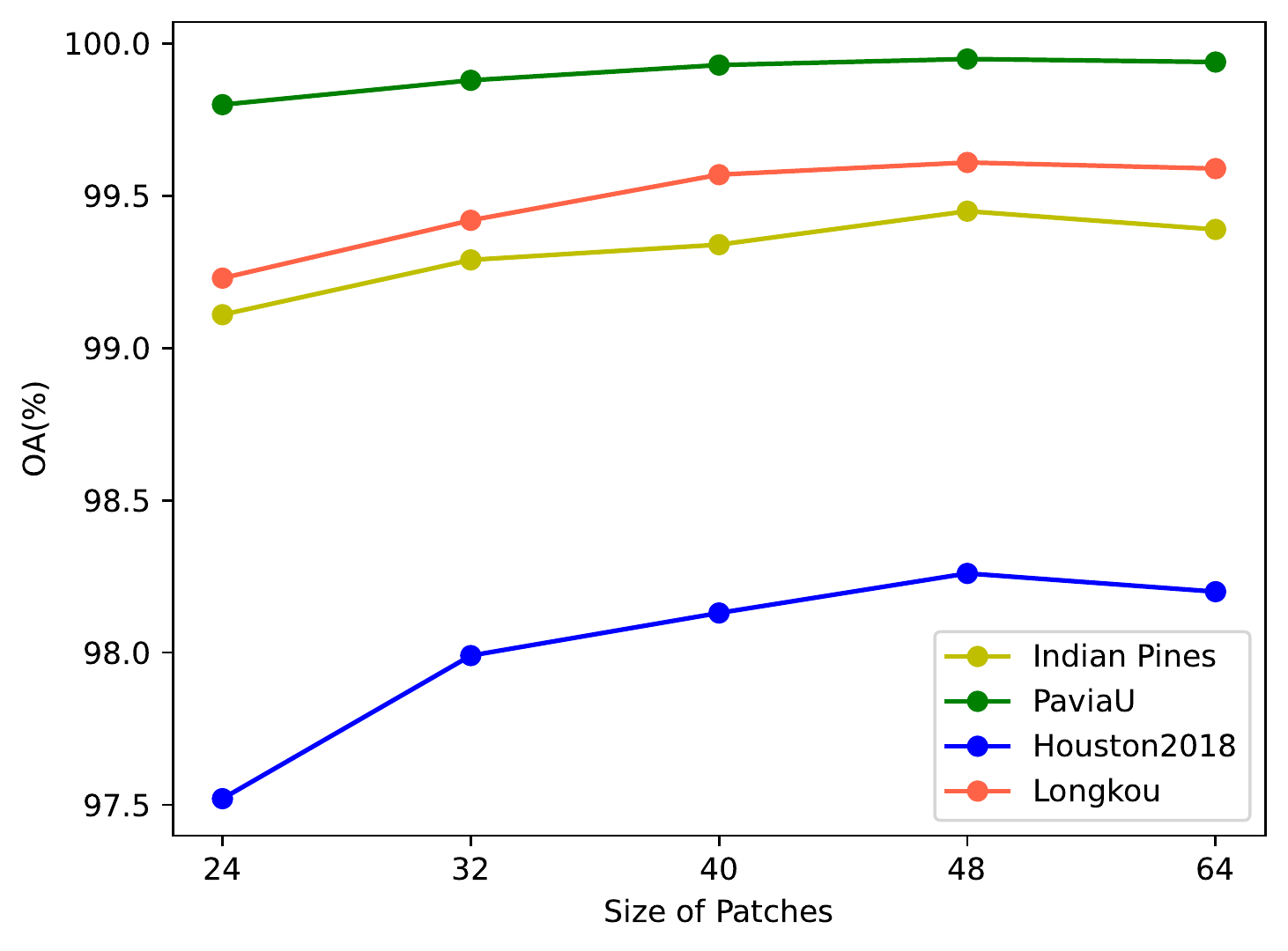}}
    \subfigure[\label{fig:results_with_pca}]{
		\includegraphics[width=0.32\textwidth]{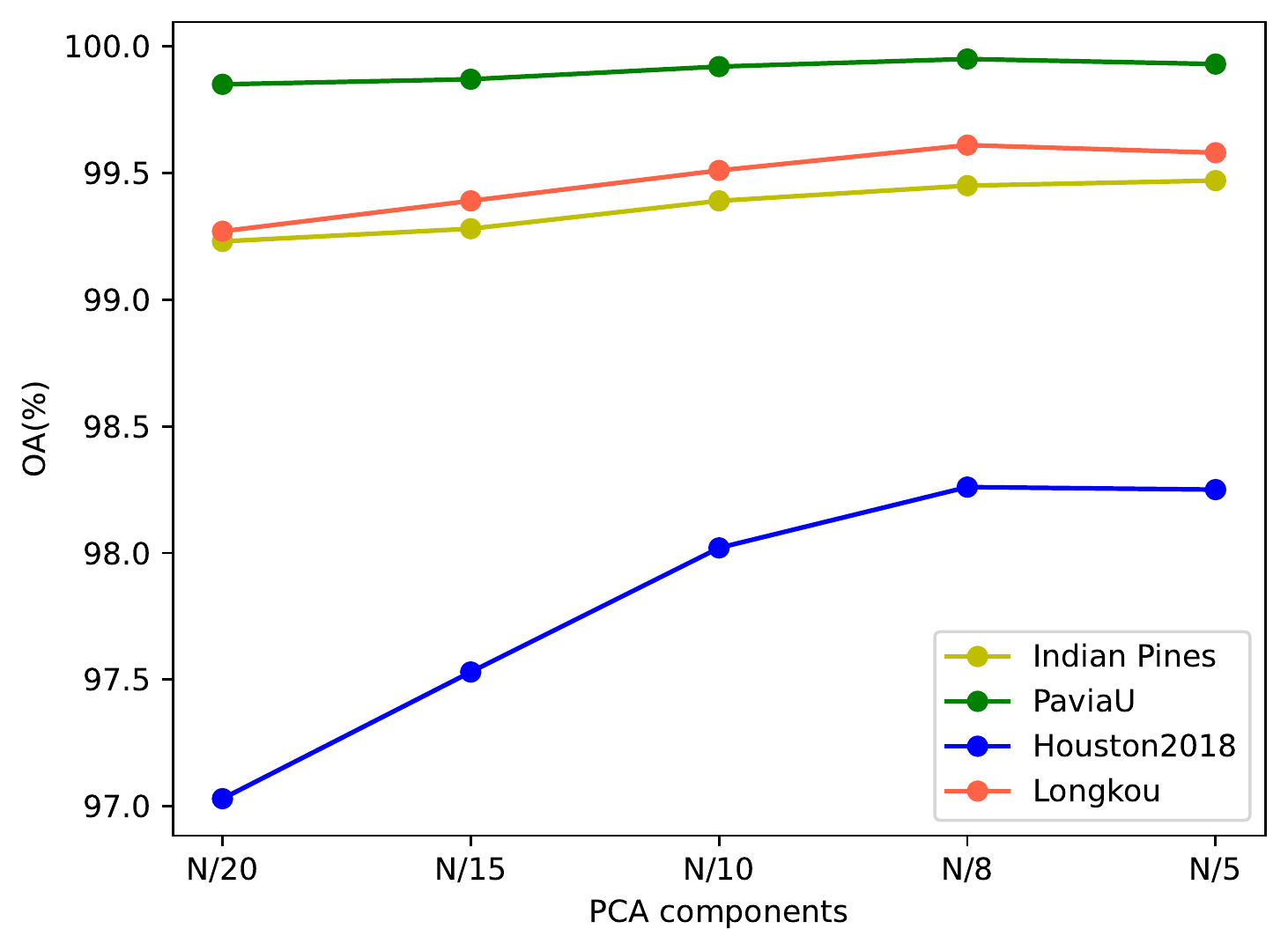}}
  \subfigure[\label{fig:results_with_steps}]{
		\includegraphics[width=0.32\textwidth]{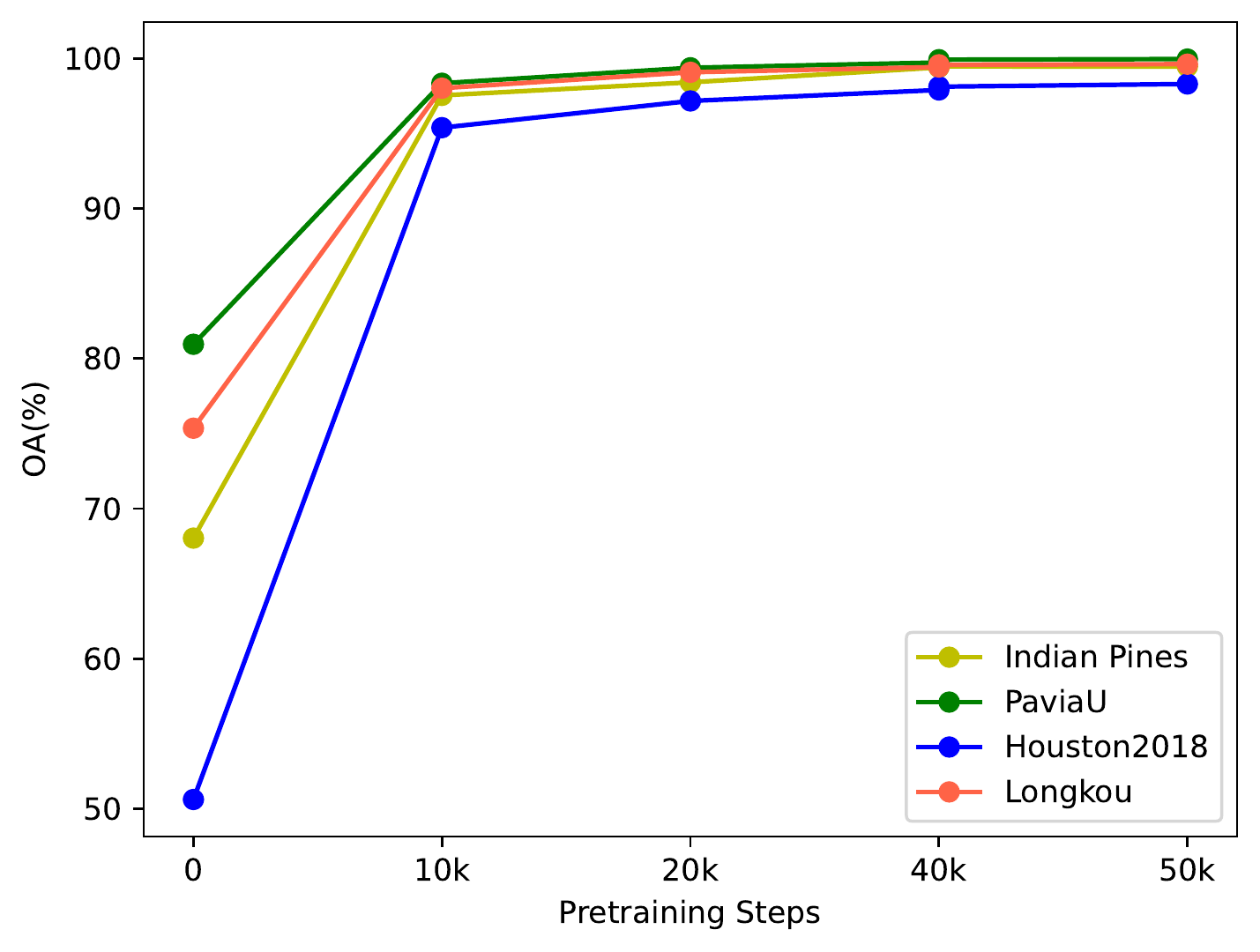}}
  
    \caption{\review{Classification results of different parameters on the Indian Pines, PaviaU, Houston 2018, and Longkou Dataset. (a) Patch size. (b) PCA component. (c) Pretraining Steps.}}
    \vspace{-1em}
    \label{fig:parameter_analysis}
\end{figure*}

\subsubsection{Visualization of Multi-Timestep Multi-Stage Diffusion Feature Exploration}
The feature map of retained and removed channels after class \& timestep-oriented multi-stage purification are visualized in Fig. \ref{fig:visualization_1}. The retained channels capture more semantic information compared to the removed ones after purification, as evident from Fig. \ref{fig:visualization_1}. Additionally, for the same channel, features at different timesteps exhibit diversity. Features at shallow timesteps tend to capture stochastic details, while those at larger timesteps focus more on higher-level semantic information.

To further demonstrate the selective timestep feature fusion’s ability to select features at different timesteps, we visualized the features with higher and lower selection weights for the same sample. As shown in Fig.~\ref{fig:visualization_2}, the results indicate that for features with higher selection weights, the response of the target pixels is more consistent with surrounding pixels of the same class, containing more classification-correlated information. If the features across all timesteps are averaged without discrimination, the important information relevant to the classification task may become obfuscated. Thus, to obtain an effective representation suitable for hyperspectral image classification,  our proposed selective timestep feature fusion increases the proportion of classification-related information in the representation by assigning higher weights to more discriminative features. The visualization provides evidence that our proposed selective timestep feature fusion can effectively select timestep features suitable for classification tasks.

\subsection{Parameter Analysis}
\label{subsection:parameter analysis}

In this section, we analyze the effect of various parameters that influence classification performance by training our proposed MTMSD in the same experimental setting as Section IV-B with different parameters.

\subsubsection{Effect of Different Patch Size}
First, we discuss the effect of different patch sizes on classification performance by two indexes, OA and AA. As shown in Fig. \ref{fig:results_with_size}, the patch size varies from 24$\times$24 to 60$\times$60. As the patch size increases, the performance first increases and then decreases. The best performance is obtained when the patch size is 48$\times$48 with OA of 99.39\% and AA of 99.30\%. Too small patches contain insufficient spatial information and too large patches reduce the attention to detailed structures. Thus, we choose 48$\times$48 to be the patch size for the proposed MTMSD.   

\subsubsection{Effect of Different PCA Components}
This section analyzes the influence of the number of PCA components, which determines how much spectral information is retained in the compressed data. As the number of PCA components increases, more spectral information is retained while more computational cost and more redundant information are brought. Since each dataset has a different number of channels, the range of PCA components is different for \review{four} datasets. Assuming that $N$ is the channel number of a dataset, the number of PCA components varies from $N/20$ to $N/5$. According to the results shown in Fig. \ref{fig:results_with_pca}, the best performance is achieved at PCA components of $N/8$.%, indicating that our proposed MTMSD can capture effective spectral information at a fixed proportion of PCA components on each dataset. 

\subsubsection{Effect of Different Pretraining Steps}
We validate the effectiveness of pretraining on the \review{four} datasets in terms of OA. As shown in Fig. \ref{fig:results_with_steps}, only 10k steps pretraining brings dramatic improvement (more than 30\% OA) to the final classification performance. Furthermore, as the pretraining steps increase, the performance continues to rise to the best at around 40k steps. 
\review{\subsection{Efficiency Analysis}
We evaluate the inference time of different methods on four public datasets to analyze the efficiency. All the experiments are carried out on a Nvidia RTX 3090. As shown in Table~\ref{inferencetime}, although our method is not the fastest in inference speed, it achieves the best classification performance since it utilizes effective multi-timestep multi-stage diffusion features. It is noted that our method has reduced the average inference time by 62\% across four datasets compared to another diffusion-based HSI classification method, SpectralDiff. 
%Although our method is not the fastest in inference speed, it achieves the best classification performance because it utilizes effective multi-timestep multi-stage diffusion features. Compared to another diffusion-based HSI classification method, SpectralDiff, our method has reduced the average inference time by 62\% across four datasets.
}
\begin{table}[t]
\renewcommand{\arraystretch}{1.3}
\caption{\review{GPU inference time(s) of different methods.
}}
\begin{tabular}{l|cccc}
\hline
Method       & Indian Pines & PaviaU & Houston 2018 & Longkou \\ \hline
2-D CNN      & 3.61         & 7.02   & 47.47        & 69.86   \\
3-D CNN      & 3.83         & 7.66   & 48.99        & 65.24   \\
SSRN         & 2.74         & 5.36   & 35.94        & 37.99   \\
SF           & 3.52         & 7.59   & 63.73        & 48.40   \\
SSFTT        & 3.79         & 7.28   & 31.38        & 59.32   \\
GAHT         & 2.51         & 5.25   & 44.90        & 34.19   \\
3DCAE        & 6.12         & 12.06  & 193.07       & 83.91   \\
3DAES        & 2.81         & 5.64   & 41.64        & 41.83   \\
UMSDFL       & 14.28        & 31.04  & 293.41       & 249.03  \\
SpectralDiff & 10.92        & 22.41  & 268.88       & 178.25  \\
MTMSD (Ours) & 3.52         & 12.41  & 168.52       & 61.53   \\ \hline
\end{tabular}
\label{inferencetime}
\end{table}

\section{Conclusion}
HSI contains rich spectral-spatial information and complex relations, which are critical for classification tasks. Many supervised and unsupervised deep learning methods are proposed to learn spectral-spatial features from HSI data, achieving promising results in HSI classification. Recently, diffusion models as powerful models in generation and reconstruction tasks have been applied to HSI classification in one recent work. However, the diffusion features used in the work are extracted solely from a single timestep
and a single stage of the denoising U-Net manually selected for each dataset, which limits the performance. Thus, we propose a diffusion-based feature learning framework that explores Multi-Timestep Multi-Stage Diffusion features for HSI classification for the first time, named MTMSD. Quantitative experiments on \review{four} HSI datasets demonstrate that our proposed MTMSD outperforms state-of-the-art supervised and unsupervised methods. 

%In the future, we will further improve the diffusion architecture, making it more suitable and specific for HSI classification. Besides, we desire to explore the applications of the diffusion model on other HSI tasks.

\bibliographystyle{IEEEtran}
\bibliography{ref} 
\begin{IEEEbiography}[{\includegraphics[width=1in,height=1.25in,clip,keepaspectratio]{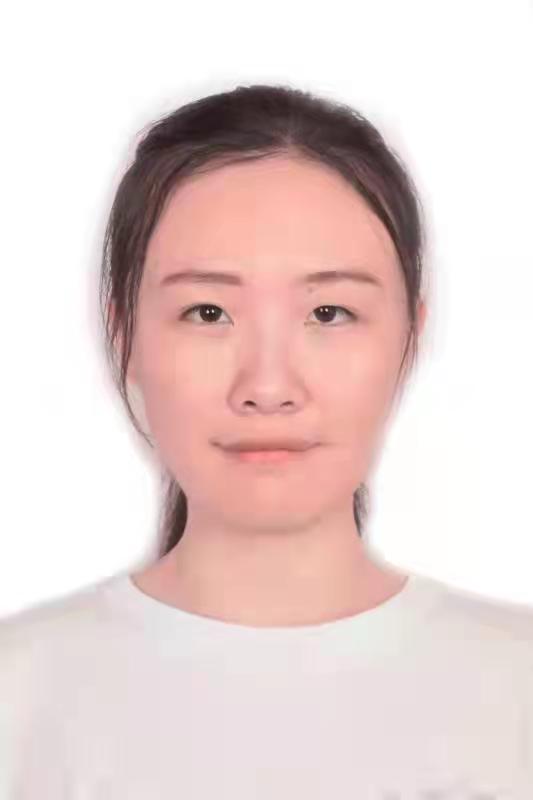}}]{Jingyi Zhou}
received the B.E. degree from the School of Information Science and Technology, Fudan University, Shanghai, China, in 2023, and she is currently pursuing the M.E. degree in the School of Information Science and Technology, Fudan University, Shanghai, China. Her main research interests include computer vision, hyperspectral analysis, sentiment analysis and depth estimation.
\end{IEEEbiography}

%\vspace{-11pt}

\begin{IEEEbiography}[{\includegraphics[width=1in,height=1.25in,clip,keepaspectratio]{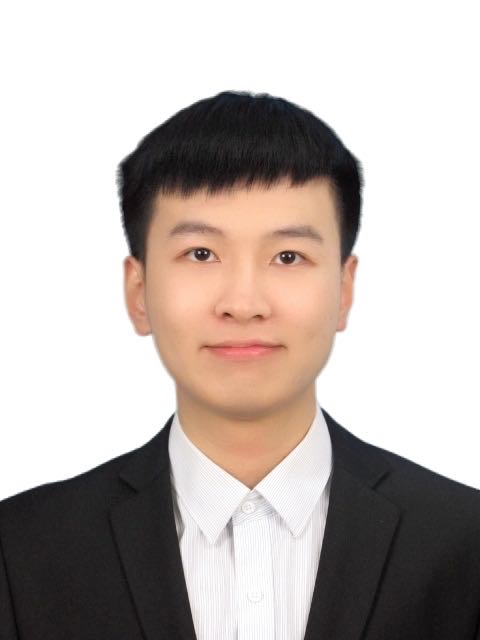}}]{Jiamu Sheng}
received the B.E. degree from the School of Information Science and Technology, Fudan University, Shanghai, China, in 2022, and now he is currently pursuing the M.E. degree in the Academy for Engineering and Technology, Fudan University, Shanghai, China. His main research interests include computer vision, image quality assessment and hyperspectral analysis.
\end{IEEEbiography}

%\vspace{-11pt}

\begin{IEEEbiography}[{\includegraphics[width=1in,height=1.25in,clip,keepaspectratio]{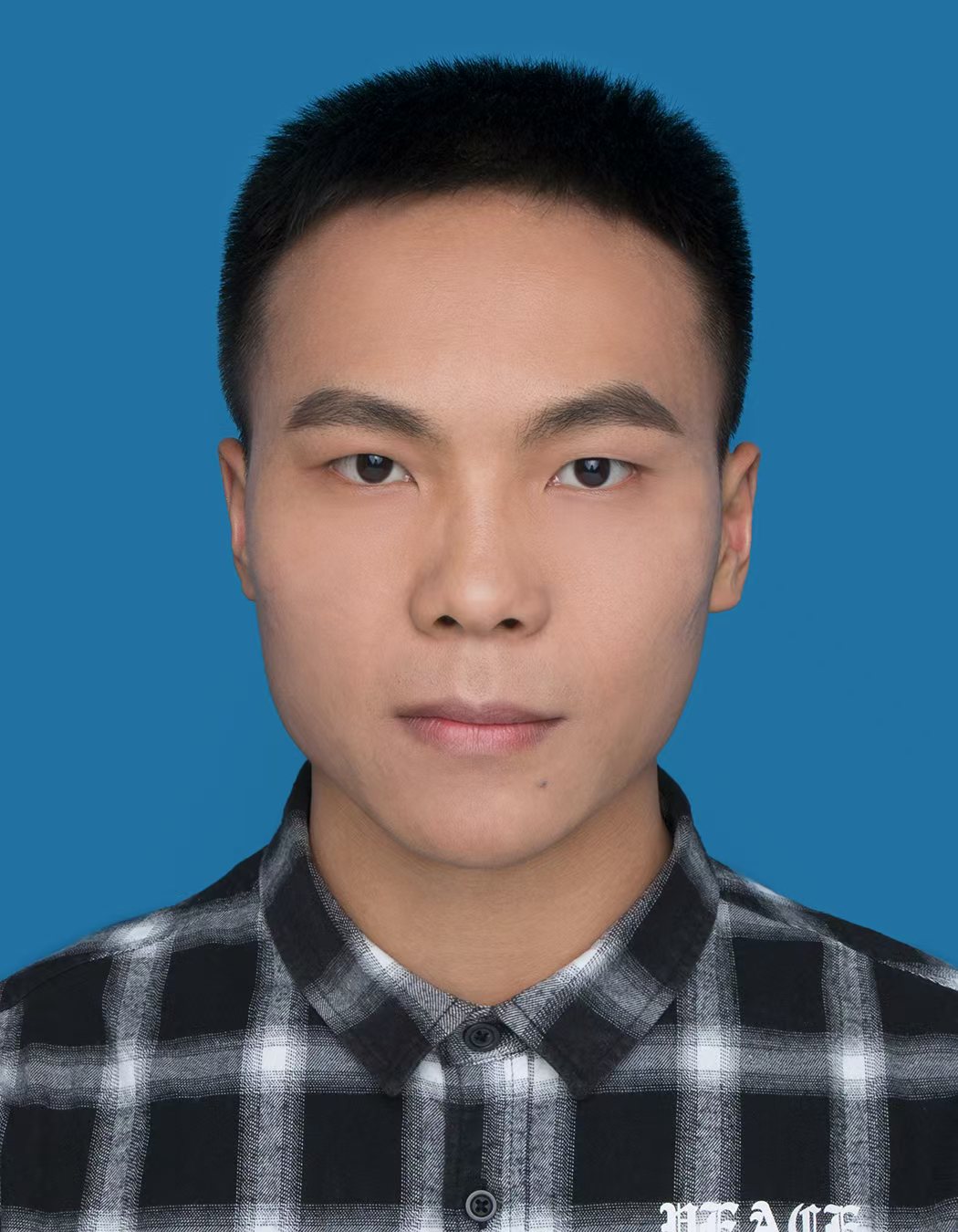}}]{Peng Ye}
is currently pursuing the Ph.D. degree with Fudan University, Shanghai, China. He has published papers in leading journals and conferences, including PAMI, IJCV, CVPR Oral, NeurIPS, ACM MM, ICME Best Student Paper, TGRS, TCSVT, and ICASSP Oral. His research interests include computer vision, network design, and network optimization. He serves as a Reviewer for various journals and conferences, including PAMI, IJCV, TCSVT, CVPR, ECCV, ICCV, and MIR.
\end{IEEEbiography}

%\vspace{-11pt}

\begin{IEEEbiography}[{\includegraphics[width=1in,height=1.25in,clip,keepaspectratio]{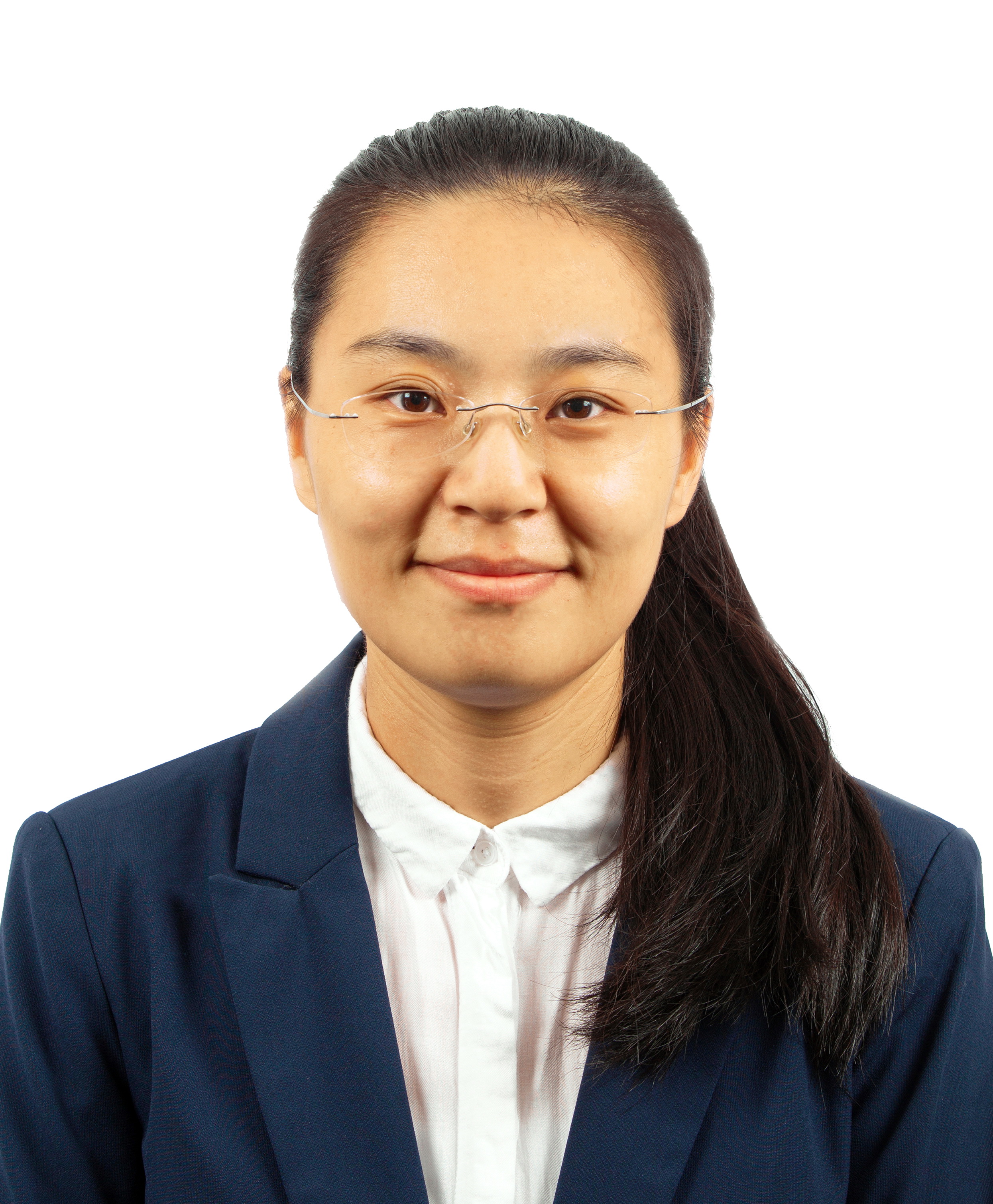}}]{Jiayuan Fan}
received the Ph.D. degree in information engineering from Nanyang Technological University, Singapore, in 2015. After her graduation, she worked as a Research Scientist at the Institute for Infocomm Research, A*STAR, Singapore. She is currently an associate professor with Academy for Engineering and Technology in Fudan University, Shanghai, China. Her main research interests include computer vision, and image forensic analysis and application.
\end{IEEEbiography}

%\vspace{-11pt}

\begin{IEEEbiography}[{\includegraphics[width=1in,height=1.25in,clip,keepaspectratio]{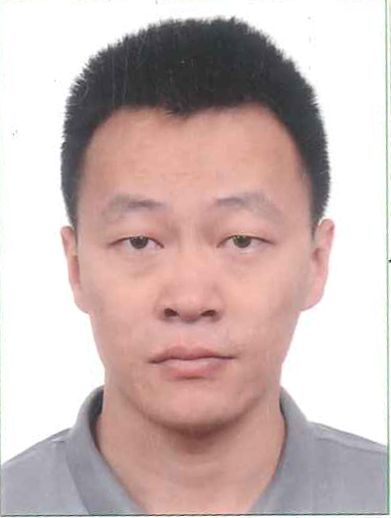}}]{Tong He}
received his Ph.D. degree in computer science from the University of Adelaide, Australia, in 2020. He is currently a researcher at Shanghai AI Laboratory. His research interests include computer vision and machine learning. 
\end{IEEEbiography}

%\vspace{-11pt}

\begin{IEEEbiography}[{\includegraphics[width=1in,height=1.25in,clip,keepaspectratio]{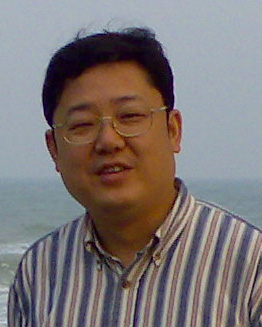}}]{Bin Wang}
received the B.S. degree in electronic engineering and the M.S. degree in communication and electronic systems from Xidian University, Xi’an, China, in 1985 and 1988, respectively, and the Ph.D. degree in system science from Kobe University, Kobe, Japan, in 1999. 
After his graduation in 1988, he was with Xidian University as a Teacher. From 1999 to 2000, he was with the Communications Research Laboratory, Ministry of Posts and Telecommunications, Kobe, Japan, as a Research Fellow, working on magnetoencephalography signal processing and its application for brain science. Then, as a Senior Supervisor, he was with the Department of Etching, Tokyo Electron AT Ltd., Tokyo, Japan, from 2000 to 2002, dealing with the development of advanced process control systems for etching semiconductor equipment. Since September 2002, he has been with the Department of Electronic Engineering, Fudan University, Shanghai, China, where he is currently a full Professor and Leader of the Image and Intelligence Laboratory. He has published more than 150 scientific papers in important domestic and international periodicals. He is the holder of several patents. His main research interests include multispectral/hyperspectral image analysis, automatic target/object detection and recognition, pattern recognition, signal detection and estimation, and machine learning.
Dr. Wang is an Associate Editor of the IEEE Journal of Selected Topics in Applied Earth Observations and Remote Sensing (IEEE JSTARS). 
\end{IEEEbiography}

%\vspace{-11pt}

\begin{IEEEbiography}[{\includegraphics[width=1in,height=1.25in,clip,keepaspectratio]{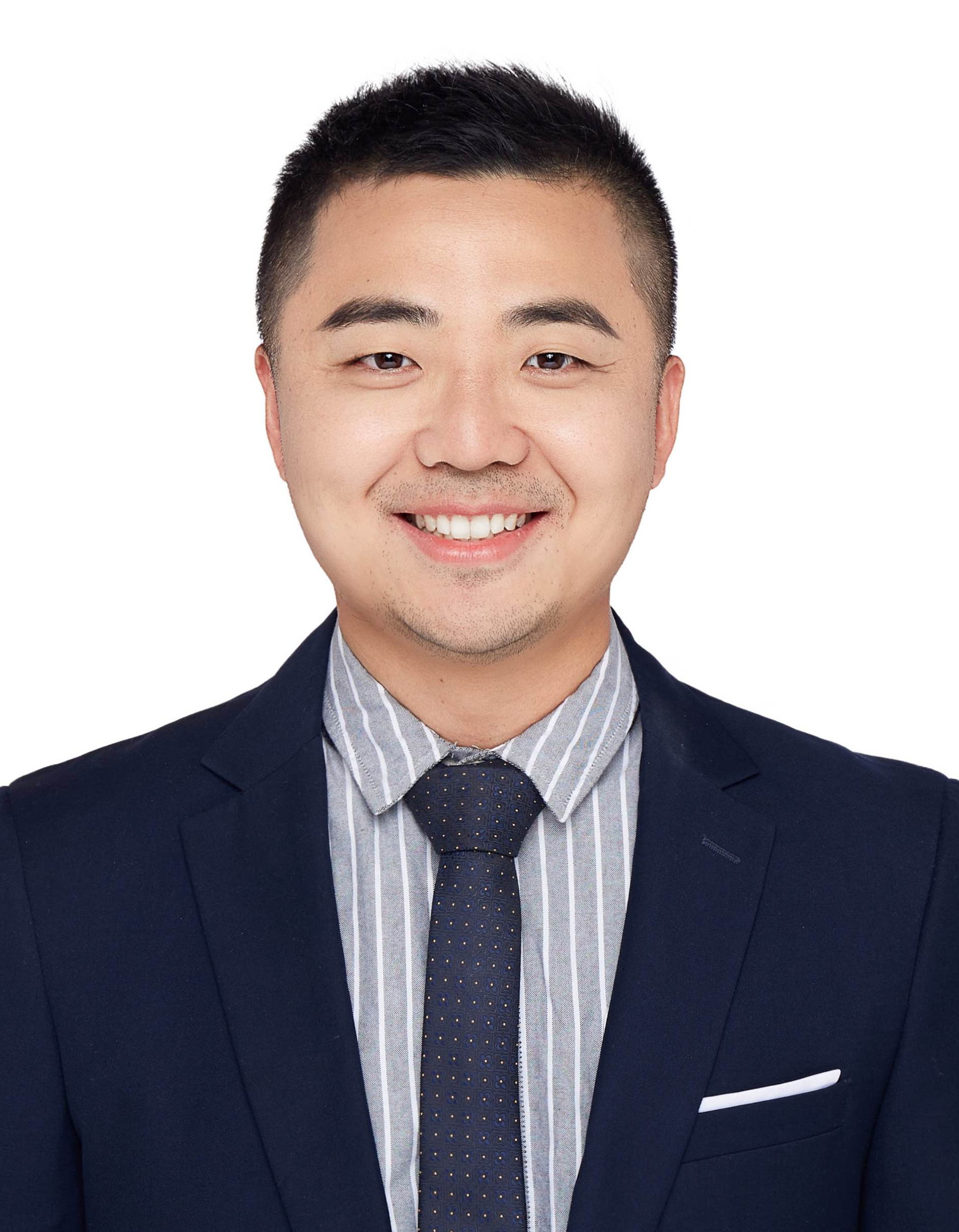}}]{Tao Chen}
received the Ph.D. degree in information engineering from Nanyang Technological University, Singapore, in 2013. He was a Research Scientist at the Institute for Infocomm Research, A*STAR, Singapore, from 2013 to 2017, and a Senior Scientist at
the Huawei Singapore Research Center from 2017 to 2018. Since 2019, he joined Fudan and led a research team focusing on light deep vision model design, multimodal vision analysis, and edge device-aware vision applications. To date, Dr. Tao Chen has undertaken multiple projects and fundings from various goverment agencies such as NSFC and corporations like Huawei, Tencent. He has published over 110 academic papers in various reputable journals and conferences like IEEE T-PAMI/IJCV/T-IP/CVPR/NeurIPS, etc., and has granted over 10 PCT patents.
\end{IEEEbiography}



\vfill

\end{document}